%% file: neurips_2026.tex
\documentclass{article}

 \usepackage[main, final]{neurips_2026}

\usepackage[utf8]{inputenc} 
\usepackage[T1]{fontenc}    
\usepackage{hyperref}       
\usepackage{url}            
\usepackage{booktabs}       
\usepackage{amsfonts}       
\usepackage{nicefrac}       
\usepackage{microtype}      
\usepackage{xcolor}         
\usepackage{enumitem}
\usepackage{amsmath}
 \usepackage{graphicx}
 \usepackage[table]{xcolor}
 \usepackage[font=small, labelfont=bf]{caption}
 \usepackage{parskip}
 \usepackage{rotating}
\title{Predicting the Needle in a Petabyte Scale Haystack: Open-Vocabulary Event Anticipation in Satellite Imagery}

\usepackage{xcolor}

\author{
Shreelekha Revankar$^{1,2}${\thanks{Corresponding Email: revankar@cs.cornell.edu}}, Mikhail Klassen$^{1}$, Creon Levit$^{1}$, Ash Hoover$^{1}$, Kavita Bala$^{2}$, Bharath Hariharan$^{2}$\\
\\
$^{1}$Planet Labs
$^{2}$Cornell University
}

\begin{document}

\maketitle

\begin{abstract}
    Satellite constellations now image the Earth daily, capturing the lifecycle of events from deforestation to urban construction. Anticipating these events before completion enables timely intervention, yet existing systems cannot jointly identify \textit{where} a change occurs, \textit{what} it will become, and \textit{when} it will finish across an open vocabulary. Building such a model requires diverse event data, but events are like needles in a haystack, making manual annotation at global scale infeasible. We address this problem by representing change as the vector difference between vision-language model embeddings at distinct time points. This arithmetic approach allows us to search for semantic transformations directly in the latent space, enabling (1) an automated data engine for global event discovery without manual annotation and (2) the largest global satellite event dataset to our knowledge \textit{SatEvents-21k} comprising 21,000+ locations across 64+ event types with~2M images from PlanetScope and Sentinel-2 spanning six continents, to train (3) a novel global-scale event anticipation model. Our model detects change with a 93.5\% max F1 score, outperforms baselines 3.5$\times$ on predicted event retrieval, and forecasts completion dates with a 3-day median error, even nine months before completion. \url{https://github.com/ShreelekhaR/predicting-the-needle}

\end{abstract}

\input{sections/new_intro2}
\input{sections/Related_Works}
\input{sections/method_3}

\input{sections/Experiments}
\input{sections/Conclusion}

\paragraph{Acknowledgments}
This work was fully funded by Planet Labs. We thank Kiruthika Devaraj,
Leela Prabhu, Megan Van Welie, and Michael Thomas for their support.

\bibliographystyle{abbrvnat}
\bibliography{bibliography}

\input{sections/Appendix}


\newpage
\input{checklist.tex}

\end{document}

%% file: sections/new_intro2.tex
\begin{figure*}[th]
    \centering
    \includegraphics[width=0.9\textwidth]{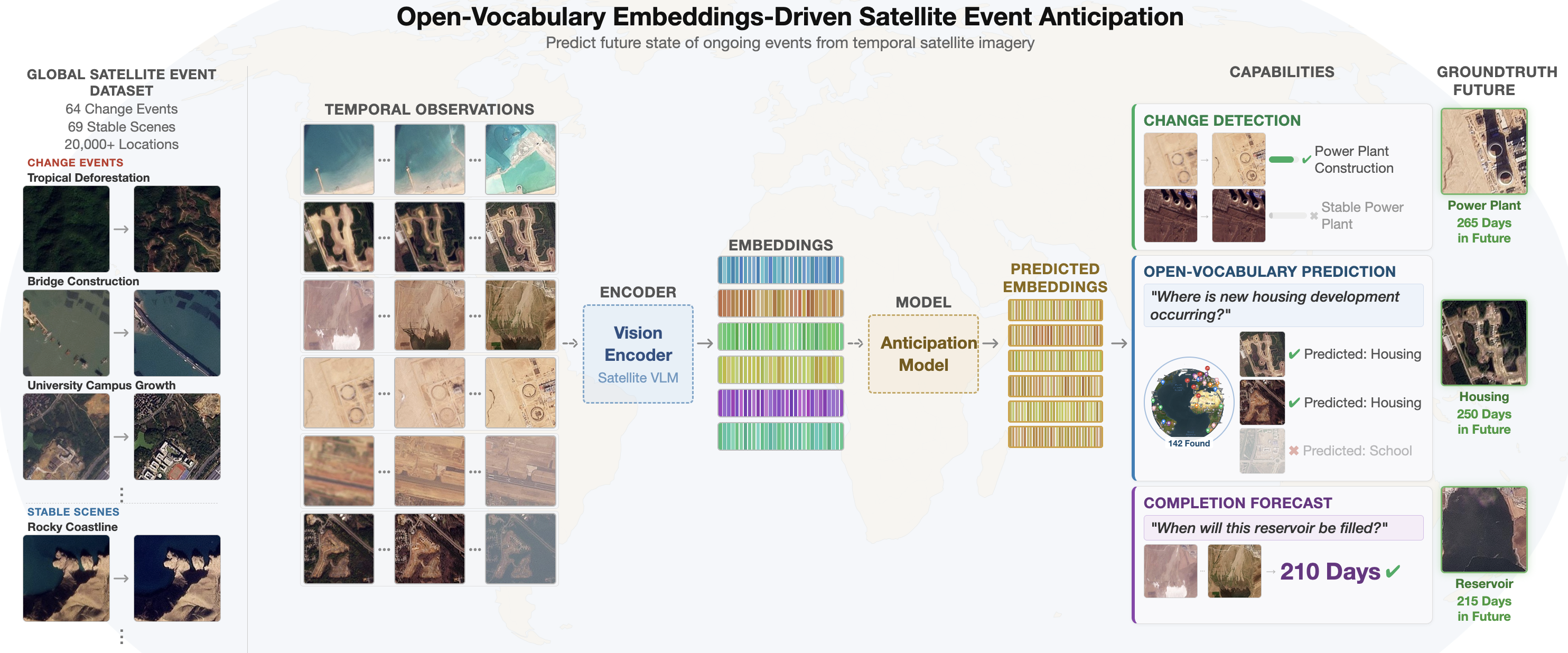}
    \caption{\textbf{We present our latent-space world model.} Trained on a large diverse set of global events, our model allows us to detect change events and predict their future states and completion dates.}
    \label{fig:title_fig}
\end{figure*}

\section{Introduction}


Satellite constellations now image the entire planet at a near-daily cadence, with public and commercial archives spanning nearly a decade of petabyte-scale records~\citep{planet, sentinel}. These archive capture the full lifecycle of events, from urban development to deforestation. Anticipating these events prior to completion enables proactive intervention, such as alerting conservation organizations months before forest canopy is lost, or informing energy planners before a data center is constructed. Building a model capable of this anticipation requires answering three questions: \emph{where} is a change event occurring globally, \emph{what} will the completed state look like, and \emph{when} will it complete, preferably in an open-vocabulary manner, without restricting to a fixed set of events or requiring task-specific annotations.

Currently no existing system satisfies all three goals, and the core challenge is data. Events we truly care about are extraordinarily rare. In a petabyte-scale archive, the majority of locations show stable, unchanging terrain. Training a model on random samples from the archive of satellite images would simply teach it to predict that nothing changes, because for most locations, nothing significant does. Manually curating annotated events at global scale is not feasible~\citep{bergman1997progressive, bell2025earth}. For this reason existing multitemporal datasets and the models trained on them remain limited in diversity and geographic coverage~\citep{revankar2025monitrs, rsicc}. More general embedding-based change detection can flag that \emph{something} changed~\citep{klassen2025igarss}, but carries no semantic signal about \emph{what} is changing or \emph{where} it is headed. 




To address these challenges, we propose an automated methodology to construct a diverse global event dataset without manual annotation. We leverage the fact that although we do not have open-vocabulary event detection models, we do have vision-language models (VLMs) that excel at recognizing \emph{static} concepts in satellite imagery \citep{barzilai2025recipe, xiong2025dofa,li2023rs}. Our key insight is that we can detect and retrieve events by recognizing the ``before" and ``after". Concretely, similar to semantic vector arithmetic in other domains \citep{radford2021learningCLIP, mikolov2013efficient}, we represent change as the vector difference between embeddings at two time points, where the latent space's semantic structure replaces manual labels. For example, the difference between an \textit{empty lot} and a \textit{data center} captures the construction event itself.



With this dataset in hand, we draw on latent-space forecasting~\citep{vondrick2016anticipating, lecun2022jepa, assran2025vjepa}, which has shown that future states can be predicted directly in embedding space with robustness to perceptual noise, a particularly valuable property for satellite imagery, which is inherently noisy~\citep{komurcu2026sat}. We build a satellite event world model that operates entirely within embedding space, ingesting a sequence of observations as a trajectory and predicting where it is headed and when it will complete. The VLM embedding space provides semantic grounding, enabling the model to answer arbitrary natural language queries about change without any task-specific fine-tuning.

In sum, our contributions are the following:
\begin{enumerate}[noitemsep, topsep=0pt, leftmargin=*]
\item An \textbf{automated data engine} that discovers any described event at a global scale without manual annotation or external data sources, leveraging VLM embedding arithmetic to localize events.
\item \textbf{The largest global satellite event dataset} to our knowledge: $\sim$21,000+ locations across 64+ event types with $\sim$2M images from PlanetScope and Sentinel-2 spanning six continents. To be publicly released.
\item\textbf{A global-scale event anticipation model} trained on temporal observation sequences that detects change with 93.5\% F1, outperforms baselines 3.6$\times$ on predicted event retrieval, and forecasts completion dates with a median 3-day error — up to 9 months before an event completes.
\end{enumerate}

%% file: sections/Related_Works.tex
\section{Related Works}

\paragraph{Retrieval and Event Detection in Satellite Imagery}
Historically, satellite imagery retrieval has faced a ``needle in a haystack'' challenge~\citep{bergman1997progressive}. Various retrieval techniques were developed~\citep{bell2025earth}, but VLMs adapted for remote sensing~\citep{liu2024remoteclip,barzilai2025recipe,mall2023remote,li2023rs,xiong2025dofa} emerged as the popular solution. While most operate over static imagery, making change characterization difficult, a small number of spatiotemporal VLMs have been developed~\citep{Jain2025TimeSenCLIPAT, Li2024UniRSUM}. Training depends on multitemporal datasets. Existing datasets often suffer from geographic constraints, limited classes~\citep{mall2022change,chen2020spatial}, reliance on manual annotation, or dependence on external sources like social media~\citep{ahmad2019social}, databases~\citep{uzkent2019learning}, or news articles~\citep{revankar2025monitrs}. Our work differs in that it requires no manual annotation, external sources, or geographic constraints, scales to arbitrary event types worldwide, and uniquely treats spatiotemporal change itself as a vector in a language-aligned latent space, queryable for open-vocabulary event retrieval.

\paragraph{Event Anticipation and World Models.}While forecasting models like TAMMs~\citep{guo2026tamms} and Sat-JEPA-Diff~\citep{komurcu2026sat} explore future state prediction, they often rely on noisy, pixel-level visual reconstruction or fixed-duration and fixed-category forecasting~\citep{irvin2024teochat,porta2025canadafiresat,komurcu2026sat}. Predicting future visual representations directly in feature space was proposed early on by~\citet{vondrick2016anticipating} to avoid the computational and noise-related pitfalls of pixel-level forecasting. A growing body of work has since explored prediction in latent space across vision and language modalities~\citep{bain2021frozen,lecun2022jepa,jiang2024vlm2vec,assran2025vjepa,chen2025vljepa,zhang2025can,zhong2025time,zou2026unlocking}. Most recently,~\citet{assran2023self} and~\citet{chen2025vljepa} have demonstrated that models can learn high-level world dynamics by predicting masked or future embeddings.~\citet{komurcu2026sat} corroborate that this latent-only approach is particularly robust to the perceptual noise inherent to satellite imagery.
However, these works largely operate on fixed event vocabularies or specific prediction horizons defined at training time. To our knowledge, none prior to us have used change vectors between scene embeddings as a  signal for open-vocabulary event anticipation across arbitrary timescales.

%% file: sections/method_3.tex
\section{A Large-Scale Satellite Event Dataset: \textit{SatEvents-21k}}
\label{sec:dataset}
To train a model capable of predicting future events from satellite imagery, we need a training dataset that captures the full diversity of events that may be of interest.
Building such a training dataset is challenging, because in most locations of the planet at most times, nothing is changing.
Therefore, if we simply sample locations and times uniformly at random, our dataset will be dominated by no change and many important events of interest will be missed.

We propose instead to start with a long list of diverse events, and retrieve locations and times where these events have occurred. To do so we leverage LLMs, using them to create a list of events, including text descriptions of the ``before'' and ``after'' states. This is done without any imagery and results in a list of events that \textit{should be} visible through satellite imagery.

Unfortunately, retrieving locations and times where a particular kind of event has occurred is challenging because we do not have open-vocabulary event retrieval models with sufficiently diverse vocabularies.
Instead, we propose to leverage satellite image VLMs that  recognize \emph{static} concepts~\citep{liu2024remoteclip,barzilai2025recipe}.
Concretely, given an event description such as \textit{Solar Farm Installation}, we use an LLM to produce a description of what the location would look like before: ``\textit{Open empty field}'' and after: ``\textit{Grid of dark rectangular solar panels}''.
We can then use static VLMs to recognize the before and after descriptions. We consider two strategies to do so:

\textbf{Strategy 1: Intersection Search.} 
We retrieve locations that match both the before description at one time point and the after description at a later time point.
In particular, we construct a database containing the embeddings $e^i_{\text{before}}$ and $e^i_{\text{after}}$ of each location $i$ at two different time points, $t_{\text{before}}$ and $t_{\text{after}}$.
Given before and after text descriptions $(d_{\text{before}}, d_{\text{after}})$ for an event, we query the database separately at each timepoint:
\begin{align}
\mathcal{R}_{\text{before}} &= \text{top-}k\left(\{\text{sim}(M^t(d_{\text{before}}), e^i_{\text{before}})\}_{i=1}^n\right) \\
\mathcal{R}_{\text{after}} &= \text{top-}k\left(\{\text{sim}(M^t(d_{\text{after}}), e^i_{\text{after}})\}\right)
\end{align}

where $M^t$ is the text encoder. Locations appearing in both result sets are candidates:
\begin{equation}
\mathcal{R}_1 = \mathcal{R}_{\text{before}} \cap \mathcal{R}_{\text{after}}
\end{equation}

\begin{figure*}[t]
  \centering
  \includegraphics[width=0.9\textwidth]{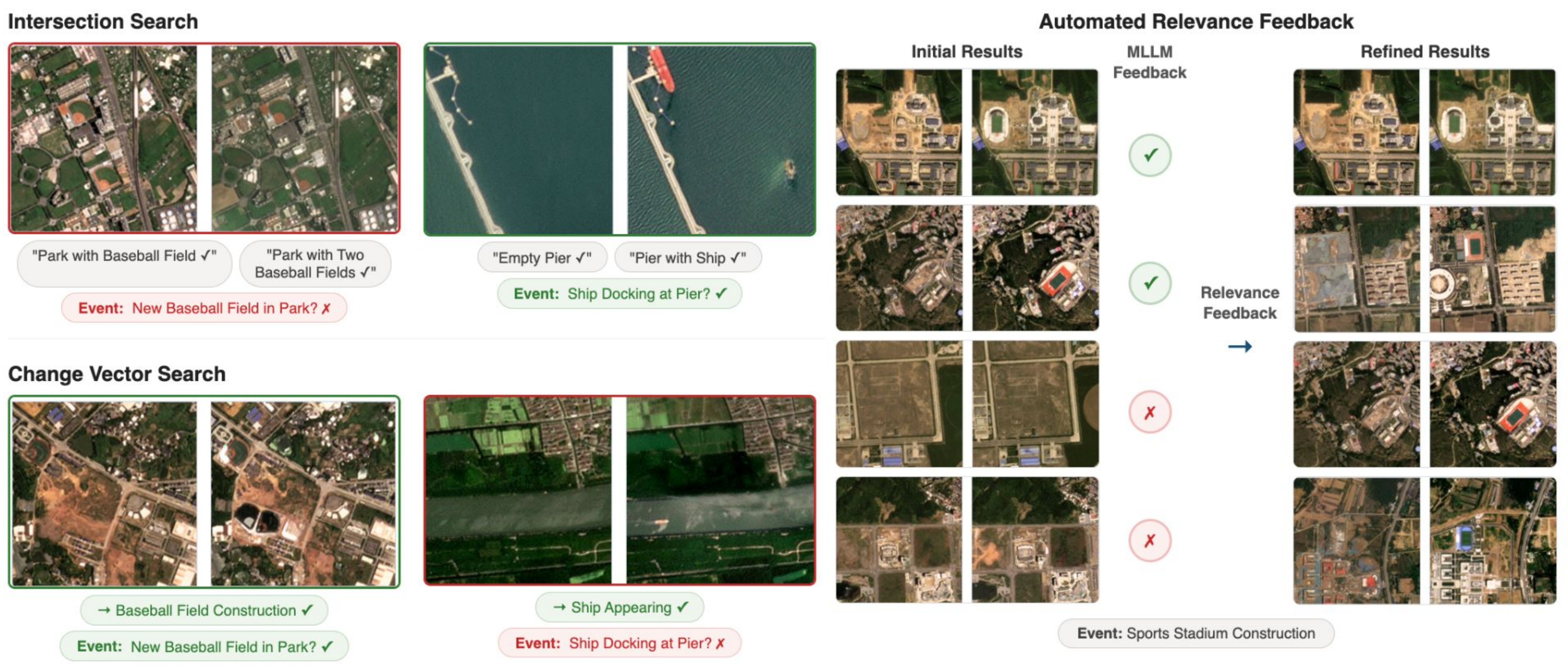}
 \caption{
    \textbf{Change-Aware Retrieval and Automated Relevance Feedback.}
    \textit{(Left)} Intersection search enforces scene context at both timepoints but can produce false positives when a scene already matches the after-state without change. Change vector search finds the correct transformation but often loses context. \textit{(Right)} ARF connects both: an MLLM scores initial intersection results, confirmed positives form averaged probe $\Delta e_{\text{probe}}$ for refined re-retrieval.
  }
  \label{fig:retrieval_and_arf}
\end{figure*}

This ensures retrieved locations both started in the expected initial state and ended in the expected final state. Thus, to retrieve locations where a ship entered the pier, we intersect a search for ``\textit{empty pier}'' with ``\textit{pier with ship}'' as shown in Figure~\ref{fig:retrieval_and_arf}.

This \textit{intersection search} has a failure mode: when the same scene type appears in both descriptions, it may retrieve locations matching both without any change occurring. For example, querying ``\textit{park with baseball field}''$\rightarrow$``\textit{park with two baseball fields}'' may surface a park with two fields at both timepoints, which is a false positive. We therefore consider a second strategy as well, outlined below.

\textbf{Strategy 2: Direct Change Vector Search.} Alternatively, we can mathematically represent the expected change using vector arithmetic:
\begin{equation}
\Delta e_{\text{query}} = M^t(d_{\text{after}}) - M^t(d_{\text{before}})
\end{equation}

This change vector encodes the semantic transformation. We query the database for similar changes:
\begin{equation}
\mathcal{R}_2 = \text{top-}k\left(\{\text{sim}(\Delta e_{\text{query}}, \Delta e^i)\}_{i=1}^n\right)
\end{equation}

where $\Delta e^i = e^i_{\text{after}} - e^i_{\text{before}}$ is the change in embedding space at the $i$-th location between the before and after time points. This retrieves locations exhibiting similar semantic transformations regardless of specific visual appearance.

This change-vector search is more discriminative, but loses scene-level context: e.g., a ``\emph{ship appearing}" vector retrieves ships anywhere, not only at piers. 

Thus both strategies have their flaws. We therefore propose to combine the two strategies using the idea of relevance feedback.


\paragraph{Automated Relevance Feedback (ARF)}

We combine the complementary strengths of the two strategies above as follows.
We use intersection search to produce a list of candidates, and then use a commercial MLLM to identify confirmed positive examples $\mathcal{P}$ from the top-\textit{k} intersection candidates.
We then compute a mean change probe following the Rocchio algorithm~\citep{rocchio1971relevance}:
\begin{equation}
\Delta e_{\text{probe}} = \tfrac{1}{|\mathcal{P}|} \sum_{i \in \mathcal{P}} \Delta e^i
\end{equation}
We then use this averaged change vector to perform change vector search (instead of constructing $\Delta e_{\text{query}}$ from text descriptions alone).
This averaged change vector captures the common semantic change pattern across confirmed positives, giving us a refined $\Delta e_{\text{probe}}$ for re-retrieval.

This iterative refinement of retrieval probes using confirmed positive examples is known as \textit{relevance feedback}~\citep{rocchio1971relevance}, which has more recently been shown effective for vision-language retrieval~\citep{khaertdinov2026little}.

\paragraph{Dataset Construction}
With these retrieval strategies and ARF, our full pipeline for retrieving events at scale is as follows:
\begin{enumerate}[noitemsep, topsep=0pt, leftmargin=*]
    \item \textbf{Generate event descriptions:} MLLM generates event types with before/after descriptions $(d_{\text{before}}, d_{\text{after}})$
    \item \textbf{Intersection retrieval:} Perform intersection search for top-$k$ candidates
    \item \textbf{ARF round 1:} MLLM identifies positive examples, compute $\Delta e_{\text{probe}}$
    \item \textbf{Change vector retrieval:} Re-query using $\Delta e_{\text{probe}}$ for refined candidates
    \item \textbf{ARF round 2:} MLLM identifies final positive examples, save to dataset
\end{enumerate}

This pipeline enables automated discovery across thousands of positive event types without manual annotation. We also apply this process to discover negative examples, i.e. stable locations where no significant change occurs, by setting $d_{\text{before}} = d_{\text{after}}$ (e.g., ``dense forest'' for both timepoints). This produces a balanced dataset of ``positive" events and ``negative" static scenes. We capture 26,039  locations spanning 
64 event classes and 69 stable scene categories across 6 continents, filtered to 21,362 locations with a sufficient number of observations.

\paragraph{Event Completion Detection}
\label{subsubsec:event_completion_detection}



Our retrieval pipeline discovers locations that changed between two timepoints. However, many applications would require us to know or predict \textit{when} the change completes. We define completion as the date beyond which anticipation provides no additional value, i.e. when the location can be retrieved reliably using its "after" state description. We devise an method to compute exactly that.

For each candidate location, we download and embed imagery between our previously retrieved two timepoints (which in our case is 1 year). Then we compute a differential change similarity signal:
\begin{equation}
S_{\Delta}(t) = \text{sim}(e_t, e_{\text{after}}) - \text{sim}(e_t, e_{\text{before}})
\end{equation}
where $\text{sim}(\cdot, \cdot)$ denotes cosine similarity, $e_t$ is the embedding at time $t$, and $e_{\text{before}}$, $e_{\text{after}}$ are the retrieved \textit{before} and \textit{after} embeddings from our event retrieval process. The metric $S_{\Delta}(t)$ is negative when the scene resembles the initial state and transitions to positive as the event completes. The zero-crossing of $S_{\Delta}$ indicates when the location appears more similar to the completed state than the initial state. Critically, this requires no additional supervision beyond the embeddings between the two retrieval timepoints. We visualize several examples in Figure~\ref{fig:completion_detection}, where the $S_{\Delta}$ signal exhibits sharp zero-crossings across diverse event types, confirming that our completion dates capture genuine perceptual transitions. Full implementation details and visualizations are in Appendix~\ref{appendix:completion_detection}.

\paragraph{Human Verification.} To quantify our data pipeline reliability and produce a human-certified test set, annotators reviewed georeferenced before-and-after image pairs and voted \textit{correct}, \textit{incorrect}, or \textit{unsure}. Classes where fewer than a majority of samples received a \textit{correct} vote were removed or re-prompted. The passing samples form our verified positive test set. Full details in Appendix~\ref{appendix:human_verification_app}. 

\section{A Model for Anticipating Satellite Events}
\label{subsec:problem_setup}




With our dataset in hand, we train a satellite event world model to predict
the future at any location. Since multi-temporal satellite sequences labeled
with open-vocabulary events are scarce, while static satellite VLMs are
trained on abundant supervision and have a rich semantic space, we predict in
the embedding space of a frozen VLM~\citep{barzilai2025recipe}. Because
successive embeddings share a spatial coordinate, the trajectory between them
represents the physical transformation at that location, and our model learns
where that trajectory is heading.

We train a model $M^{\text{pred}}$ that takes in a sequence of embeddings $\{e_0, \ldots, e_n\}$ as well as a time horizon $h$ and predicts the embedding at that location $h$ days from the last observation:

\begin{equation}
e_{\text{pred}}(h) = M^{\text{pred}}(\{e_0, \ldots, e_n\},\, h)
\end{equation}

where $e_0$ is the embedding of the oldest observation, $e_n$ is the most recent, and $h \in \mathbb{R}^+$ is the queried prediction horizon in days. This formulation allows the model to produce different predictions for different horizons from the same input sequence.

\paragraph{Predicting time-to-completion}
In addition to predicting what a location would look like at a given time in the future, for locations going through events, we may also want to know when the event will complete. 
To provide this additional capability, we also add an additional head to our model that predicts the time-to-completion, $t_{\text{pred-comp}}$.
Thus, our model predicts both the future embedding at an input time horizon, as well as the predicted time to completion for any ongoing events:
\begin{equation}
\label{equation:model_objective}
e_{\text{pred}}(h), t_{\text{pred-comp}} = M^{\text{pred}}(\{e_0, \ldots, e_n\},\, h)
\end{equation}

\subsection{Model architecture}
Our model $M^{\text{pred}}$ is a causal transformer that ingests the full observation sequence prepended with a learnable before-state token: $[\texttt{BEFORE},\, e_0, \ldots, e_n]$. Each token is augmented with temporal position encodings derived from its observation date. The transformer processes this sequence with causal self-attention, producing a context-aware representation of the location's temporal trajectory.

The queried future horizon $h$ (in days) is encoded via a sinusoidal temporal encoding into a query token. This query token then attends to the encoded sequence through a cross-attention layer, producing a horizon-conditioned representation from which the final prediction $e_{\text{pred}}$ is decoded via a feedforward head. The architecture supports arbitrary prediction horizons.

To make temporal dynamics explicit, we augment each embedding with its forward \textit{velocity}---the change to the next observation:
\begin{equation}
\tilde{e}_i = W \cdot [e_i;\, e_{i+1} - e_i] + b
\end{equation}
where $[\cdot;\cdot]$ denotes concatenation and $W, b$ are learned projection parameters followed by layer normalization. This is analogous to providing optical flow~\citep{horn1981determining}: rather than requiring the model to infer rates of change implicitly from consecutive embeddings, we provide the change signal directly, allowing the model to distinguish accelerating from decelerating events.

\subsection{Training Objective}
\label{subsubsec:training_objective}
We sample locations from our dataset, which includes both locations where events have occurred as well as ``negative" locations where nothing changes. For each location, we sample a sequence of time points and corresponding satellite images before the event occurred, as well as a horizon $h$ for which the model should predict the embedding.
We train $M^{\text{pred}}$ to predict the embedding at the input time horizon $h$, supervised against the actual observed embedding at that time horizon. We combine a directional prediction term (based on cosine similarity) with magnitude regularization:
\begin{equation}
\mathcal{L}_{\text{pred}} = \underbrace{\left(1 - \text{sim}(e_{\text{pred}}(h),\, e_{\text{gt}}(h))\right)}_{\text{completion state alignment}}
+ \;\lambda \cdot \underbrace{\left(\|e_{\text{pred}}(h)\| - \|e_{\text{gt}}(h)\|\right)^2}_{\text{magnitude regularization}}
\end{equation}
where $e_{\text{gt}}(h)$ is the observed embedding at time horizon $h$ and $\lambda$ is a weighting coefficient. The cosine loss encourages directional alignment between predicted and ground-truth embeddings, while the magnitude loss preserves embedding scale.

\paragraph{Time-to-Completion Head.}
In addition to predicting \textit{what} the completed state will look like, we jointly train a scalar regression head to predict \textit{when} the event will complete:
\begin{equation}
\mathcal{L}_{\text{time}} = \text{MSE}\left({t}_{\text{pred-comp}},\; t_{\text{comp}}\right)
\end{equation}
where ${t}_{\text{pred-comp}}$ is the predicted time-to-completion (in days) and $t_{\text{comp}}$ is the ground-truth lead time. This head branches from the same hidden representation as the embedding predictor, adding negligible parameters. The full training objective is:
\begin{equation}
\mathcal{L} = \mathcal{L}_{\text{pred}} + \mu \cdot \mathcal{L}_{\text{time}}
\end{equation}
with $\mu = 0.1$. The time head is trained only on positive (event) samples; negative (stable) locations contribute only to $\mathcal{L}_{\text{pred}}$. 

This allows us to directly output ${t}_{\text{pred-comp}}$, an estimate of the number of days until event completion. Combined with the known observation date, this yields an absolute predicted completion date.



\subsection{Inference -- Event anticipation}
Once trained, we can use our model for event anticipation. Concretely, given an event description, we consider the task of retrieving all geographical locations where the event is \emph{predicted to happen in the future} (in a time horizon $h$). Inference consists of two separate stages:

\paragraph{Stage 1: Anticipation.} 
For every location $i$ in our database, we predict the future embedding $h$ days later using the observations till now, obtaining $e_{\text{pred}}(h)$ and $ t_{\text{pred-comp}} $ for all locations. This stage takes \textbf{no text input}, so its outputs are query-independent and computed once per location and horizon. 

\textbf{Change detection:} Of course, when anticipating events, we only want to retrieve locations where a change is actually taking place.
We therefore pre-filter based on the predicted change vector magnitude $\|\Delta e^i_{\text{pred}}\|$. Locations where this predicted change magnitude is less than a threshold $\epsilon$ are not considered for retrieval.
\begin{equation}
\label{eq:detection_threshold}
\Delta e^i_{\text{pred}}(h) = e^i_{\text{pred}}(h) - e^i_n
\end{equation}

\paragraph{Stage 2: Retrieval.}
For any set of events, we use an LLM to produce ``after" text descriptions $d_{\text{after}}$. With these descriptions, we compute their text embeddings $M^t(d_{\text{after}})$ via the frozen VLM's text encoder. We can now score and rank each remaining location based on the cosine similarity between the predicted image embedding and query ``after" text embedding:
\begin{equation}
\label{eq:model_inference}
\mathcal{R}_{pred} = \text{top-}k\left(\{\text{sim}(M^t(d_{\text{after}})), e^i_{\text{pred}}(h)\}\right)
\end{equation}

This is a nearest-neighbor lookup in the shared VLM embedding space. Retrieving a new
event category costs only one text encoding and needs no rerun of Stage 1.

%% file: sections/Experiments.tex
\section{Experimental Setup}
\label{sec:experiments}

\input{tables/retrieval}

\begin{table*}[t]
    \centering
    \begin{minipage}{0.48\textwidth}
        \centering
        \input{tables/leadtime}
    \end{minipage}
    \hfill 
    \begin{minipage}{0.48\textwidth}
        \centering
        \input{tables/timetocompletion}
    \end{minipage}
\end{table*}

\subsection{Embedding Models}                             We use two satellite Vision-Language Models as frozen feature extractors:                  
\begin{enumerate}[noitemsep, topsep=0pt, leftmargin=*]       
\item \textbf{RS-SigLIP}~\citep{barzilai2025recipe} (1152-dim): A SoTA satellite VLM 
from Google Research Earth AI Remote Sensing Foundations suite, available to researchers.

\item \textbf{RemoteCLIP Fine-tuned}~\citep{liu2024remoteclip} (768-dim): Off-the-shelf public satellite VLMs performed poorly on our event taxonomy. We fine-tuned RemoteCLIP with LoRA for 2 epochs on before/after image-text pairs to bring it to a usable retrieval baseline(Table~\ref{tab:event_retrieval_dataset}). 

\end{enumerate}                    
Neither VLM receives sequential data, both are trained only on static image-text pairs. All temporal reasoning is learned by our anticipation model over frozen embeddings.
We additionally compare several open-source satellite VLMs on the task of event retrieval in Appendix~\ref{appendix:event_retrieval}.

\subsection{Event Generation}
We utilize Gemini 3.0 Pro~\citep{gemini_team_2023_gemini1} to generate a diverse, open-vocabulary taxonomy of satellite events. For each event, the model generates natural language descriptions of the pre- and post-conditions, which serve as the semantic targets for our latent-space arithmetic.

\subsection{Prediction Models}

\begin{itemize}[noitemsep, topsep=0pt, leftmargin=*]
    \item \textbf{Satellite Event Anticipator (Ours):} A transformer-based architecture trained to predict the future latent $e_{\text{pred}}(h)$ from partial observations $\{e_0, \ldots, e_n\}$ and a queried horizon $h$. Ablation study on backbone and additional implementation details in Appendices~\ref{appendix:architecture_ablation} and~\ref{appendix:model_implementation_details}.
     \item \textbf{TeoChat~\citep{irvin2024teochat}:} Satellite-specific multimodal LLM. 
    \item \textbf{Gemini 3 Flash~\citep{gemini_team_2023_gemini1}:} A frontier multimodal LLM. 
    \item \textbf{Sat-JEPA-Diff~\citep{komurcu2026sat}:} A diffusion-based satellite image prediction model. 
   \item \textbf{Static (No-Change):} Predicts that nothing has changed. It uses the most recent observation as the future prediction ($e_{\text{pred}} = e_n$).
    \item \textbf{Oracle:} This represents the upper-bound for prediction. It uses the true future embedding ($e_{gt}$) for locations where an event occurs, combined with perfect change detection.
\end{itemize}

For the MLLMs, Gemini and TEOchat are prompted to identify whether change is occurring, classify event type, and estimate days to completion. We embed the classification outputs to calculate retrieval metrics using RS-SigLIP text encoder. For SAT-JEPA-DIFF, we provide the most recent observation in each sequence and embed its predicted image with RS-SigLIP image encoder for retrieval comparison. Unlike our model, it accepts no temporal context, only a single input frame.

\subsection{Benchmark}                                                    
\label{subsubsec:benchmark}

\paragraph{Train/Test Splits.}
Our dataset \textit{SatEvents-21k} contains 177,241 samples from 21,362 unique locations, split by S2~\citep{s2geometry} cell ID with zero location overlap between splits. Training uses 132,908 samples (64,608 positive, 68,300 stable). Testing uses 43,533 samples: 25,368 from seen classes (11,275 positive, 14,093 stable) plus 18,965 out-of-vocabulary samples (3,187 positive from 10 randomly sampled held-out event classes) to evaluate generalization to unseen event types. Three geographic holdout regions containing 517 locations (4,510 samples) test spatial generalization. We include a map visualizing the dataset and its subsets in Figure~\ref{fig:dataset_dist}, Appendix~\ref{appendix:dataset_map}. 

\paragraph{Temporal Event Sampling.}
For each positive location, we sample at four fixed lead times before event completion: 1, 3, 6, and 9 months, with 5 additional random lead times, ensuring coverage across different stages of event progression. Each sample consists of the weekly embedding sequence up to the cutoff date, with a median of 19 input embeddings per sample with PlanetScope (PS) MCLA. We use the same locations and lead times with Sentinel-2 (S2); however, due to S2's lower revisit cadence (~5 days vs daily for PS), the median sequence length is 14 embeddings per sample. At test time, all models are evaluated on the same 43,533 samples with identical locations and lead times across both sensors, with S2 naturally exhibiting gaps for some observation periods.  

\subsection{Metrics}
We evaluate using max F1 scores for change detection (thresholding predicted change magnitude $\|e_0 - e_{\text{pred}}\|$ at $\epsilon$ $\approx$ 0.245), mean average precision (mAP@$k$) for event and predicted event retrieval (ranking test samples by cosine similarity to \textit{after} text queries, with undetected samples scored $-\infty$), and Pearson correlation $r$ for completion date estimation.

To test robustness and statistical significance of our results we perform a bootstrapping experiment by subsampling within our test set 1,000 times, and a permutation test on the per-class AP between the static baseline and our model using 10,000 permutations. We also report false-positive rate (FPR), the fraction of stable samples
whose predicted change magnitude exceeds the max-F1 threshold. We assess this per prediction horizon, and per land-cover class, and assess expected false alarms per 1000~km$^2$.

\begin{figure}[t]
    \centering
    \begin{minipage}{0.49\textwidth}
        \centering
        \includegraphics[width=\textwidth]{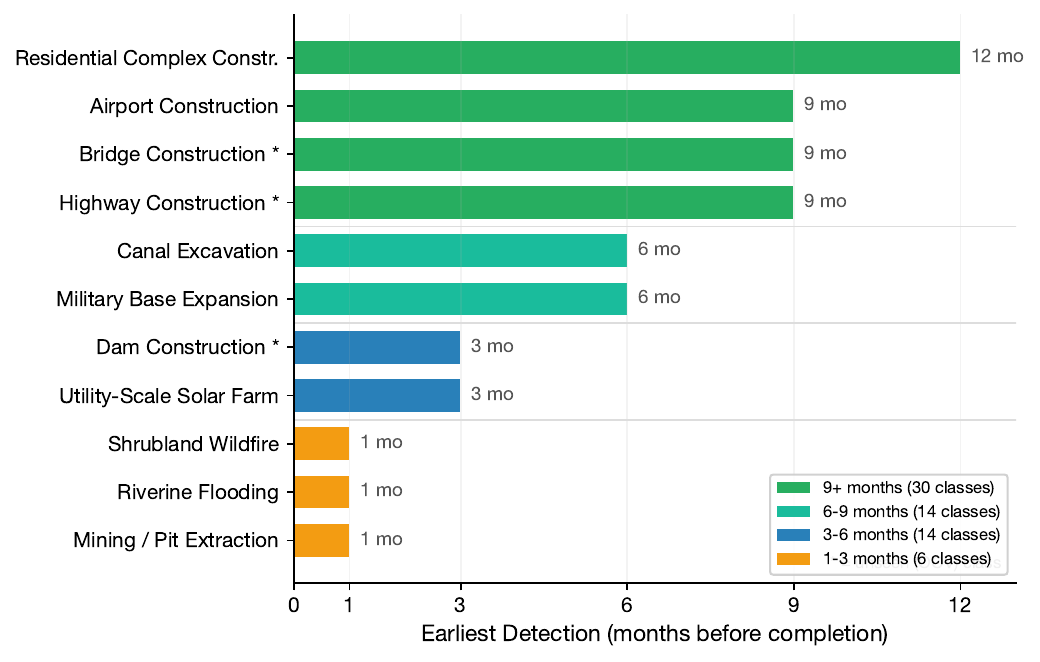}
        \caption{Earliest detection horizon per event class. * denotes classes unseen during training.}
        \label{fig:detection_horizon}
    \end{minipage}\hfill 
    \begin{minipage}{0.48\textwidth}
        \centering
        \includegraphics[width=\linewidth]{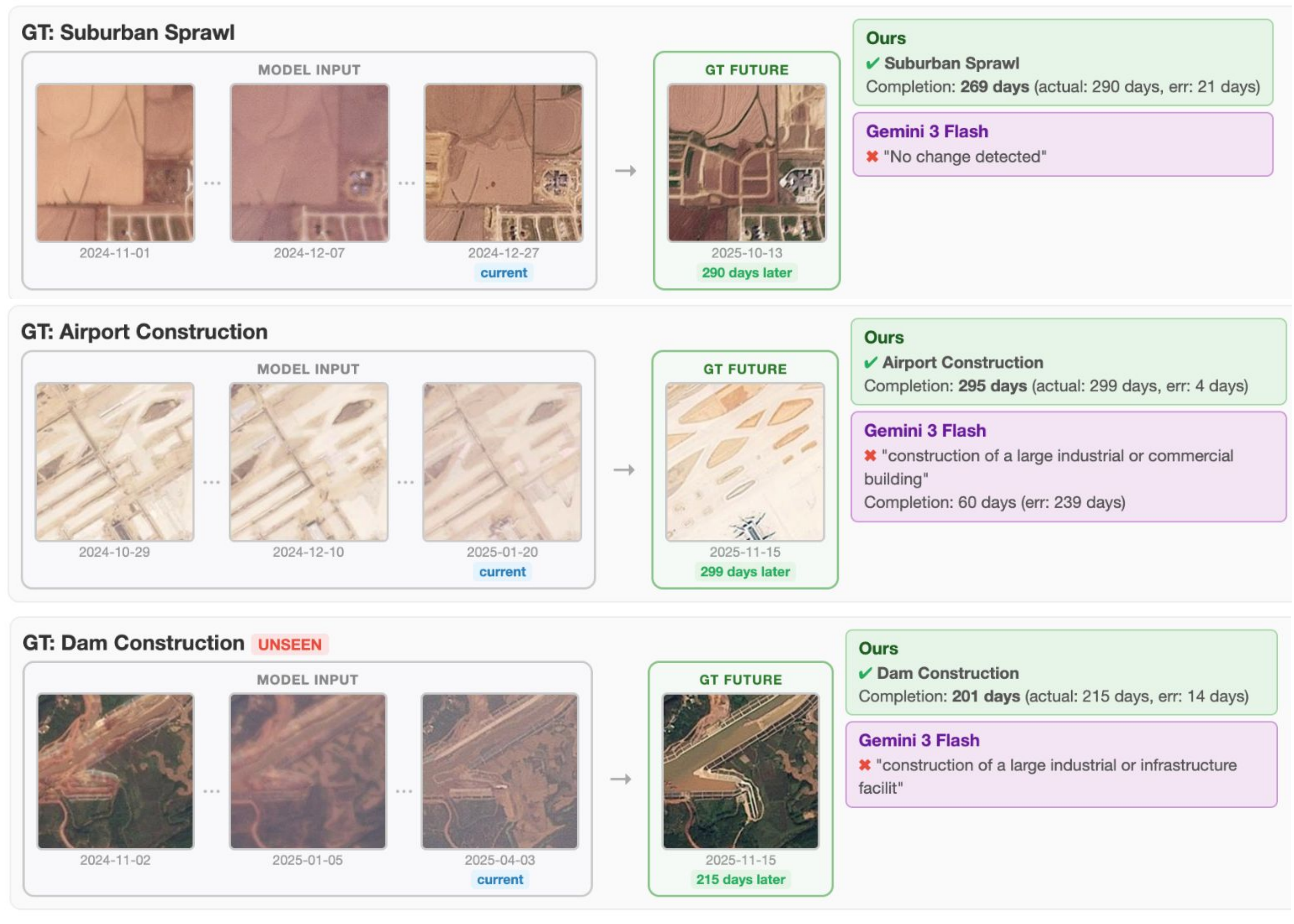}
        \caption{Qualitative examples of our model. Additional examples in Appendix~\ref{appendix:qualitative_overall}.}
        \label{fig:second}
    \end{minipage}
\end{figure}
\vspace{-0.5 em}

\section{Results}
\label{sec:results}

\paragraph{Open Vocabulary Event Detection and Anticipation Performance.}
As seen in Table~\ref{tab:Retrieval}, our best model achieves a detection F1 of 0.935 and mAP@5 of 0.533, a 3.4$\times$ improvement over the Static baseline (0.155). At global monitoring scale, the vast majority of locations are stable; our change magnitude gating achieves F1 of 0.935, effectively filtering negatives before retrieval. The model generalizes both to held-out event types (10 unseen classes) and unseen geographic regions (Geo Unseen mAP@5 of 0.429). To verify robustness, we bootstrap 1000 test set subsamples and find the 90\% confidence interval spans ±0.092 mAP across all 64 seen and unseen classes, and our permutation test of 10,000 permutations on the per-class AP between the static baseline and our model confirms statistical significance at p < 0.001. SAT-JEPA-Diff performs below the Static baseline at all horizons, demonstrating that a single input frame cannot capture the trajectory of an ongoing event. Gemini 3 Flash, despite processing the full image sequence, also underperforms, demonstrating that even frontier MLLMs struggle with the dynamics of satellite events.

\paragraph{False Positive Rates.} 

Our overall FPR is 5.56\%. In PlanetScope imagery, a 1000 km² area involves ~2,200 image chips. With our average FPR rate of 5.56\% at our reported max F1 threshold, this translates to ~123 false alarms per 1000 km². 

To assess FPR as a function of prediction horizon (time). A natural concern is that longer-horizon predictions would produce more false alarms because uncertainty compounds. To test this, we ran inference on 5,000 randomly sampled stable-scene test samples at four query offsets: 30, 90, 180, and 270 days into the future, holding the input sequence fixed. FPR is essentially invariant to horizon: 5.56\% at 30 days, 5.56\% at 90 days, 5.56\% at 180 days, and 5.58\% at 270 days, with only 1 of 5,000 samples flipping across the threshold from the shortest to the longest lookahead. Extending the anticipation horizon does not inflate the false-alarm rate on stable land, because the model was trained to predict "no change" for stable scenes regardless of the queried date.

To assess FPR as a function of stable-scene land-cover class. We computed per-class FPR on the negative samples pool. Classes with visually uniform content achieve 0.0\% FPR (Open Ocean, Stable Forest, Salt Flats, Seasonal Grassland). The highest per-class FPR is Coastal Beach at 43.2\%. We note this may partly reflect a reasonable model behavior rather than a pure error, since beaches change across tidal cycles, seasonal erosion, and storms. 

\paragraph{Forecasting Horizon and Lead Time.}
Table~\ref{tab:lead_time} analyzes retrieval across various lead times. Our model maintains a stable predictive signal across all horizons: at 6-9 months, it achieves 8.5$\times$ the performance of the static baseline. Even beyond 9 months, the model retains a meaningful signal (0.245). Both SAT-JEPA-Diff and Gemini degrade to near-zero at all horizons, performing below the Static baseline.

How early we can detect an event depends heavily on event type. As seen in Figure~\ref{fig:detection_horizon}), large-scale civil infrastructure is detectable up to 12 months before completion, while rapid-onset natural events such as wildfires and flooding allow only 1 month of anticipation. This trend holds even for novel classes unseen during training (Bridge and Highway Construction, Dam Construction).

\paragraph{Time-to-Event Accuracy.}
A robust world model should not only predict \textit{what} will happen, but \textit{when}. Our model additionally predicts event completion dates via a scalar regression head (Table~\ref{tab:time_completion}). On PlanetScope imagery, it achieves a Pearson correlation of $r{=}0.997$ with median absolute error of 3 days and 92.1\% of predictions within 14 days. On Sentinel-2, with sparser temporal coverage, it still achieves $r{=}0.992$ with median error of 6 days. 

\paragraph{MLLM Performance \& Scalability.} 
While commercial MLLMs demonstrate competitive classification accuracy on detected changes, they achieve substantially lower retrieval mAP because they produce a single text answer per location rather than a rankable embedding.
They also fail to anticipate satellite events at long horizons and cannot predict completion timelines. Practically, MLLMs also require inference with the full sequence per location per query, whereas our approach retrieves across all locations via vector similarity, making repeated queries feasible at scale.

\paragraph{Robustness to Noise.} We apply no explicit cloud-masking, atmospheric correction, or pixel-level denoising to the model's inputs. Our sequences have inconsistent gaps in observation cadence, and include visible snow cover, cloud haze, and partial occlusions. 
By operating in VLM embedding space rather than pixel space, our model avoids modeling pixel-level noise. Images with noise or sensor artifacts still produce meaningful embeddings, and within a temporal sequence, noisy frames are naturally downweighted relative to clearer observations. This robustness to low-level noise is a well-studied property of joint-embedding methods. JEPA and JEPA-family objectives have been shown to prioritize abstract, slowly-varying semantic features over unpredictable pixel-level detail~\citep{sobal2022joint,littwin2024jepa}.

\paragraph{Limitations.}
Our data engine bypasses common geographic and categorical limits but currently assumes a single dominant event between endpoints. Consequently, intermediate transitions, such as an empty lot becoming a parking area before becoming a shopping mall, are not captured. Furthermore, while this study uses a one-year timespan, the same process can scale from daily to multi-year cadences as embeddings archives grow.  Additionally we recognize that the term ``completion'' can be ambiguous. We compare our embeddings based completion detection against human completion labels in Appendix~\ref{appendix:timetocompletionambiguity}.Importantly, the world anticipation model’s performance is capped by its VLM embeddings, as concepts foreign to the VLM remain unrecognized in our label-free training. However, we show in Appendix \ref{appendix:event_retrieval} that finetuning the VLM on our pipeline’s image-caption pairs improves retrieval, suggesting our data pipeline may help mitigate the very limitations the model inherits. We also assess how well we detect, retrieve, and estimate events not collected via embeddings we created an independent unbiased test set by labeling a small dataset of logging/deforestation events recorded in publicly available state logging permits records~\cite{logging_permit}. These assessments are included in Appendix~\ref{appendix:logging_permit}.


\paragraph{Broader Impact.}
\label{para:broader_impact}

This research transitions Earth observation towards predictive intelligence. Such foresight provides a crucial window for intervention, allowing humanitarian and environmental groups to act before events fully escalate. However, the forecasting capability afforded by this model carries risks of dual-use. While our dataset is curated to exclude sensitive content, the underlying methodology could be repurposed for surveillance. To mitigate these risks, we emphasize that the scaling of such datasets must be accompanied by ethical governance frameworks that prevent the exposure of sensitive activities.




%% file: tables/retrieval.tex
\begin{table*}[t]
\centering
\resizebox{0.8\textwidth}{!}{%
\begin{tabular}{lcc|ccc|ccc|ccc|ccc}
\toprule
 & & & \multicolumn{3}{c|}{\textbf{Seen}} & \multicolumn{3}{c|}{\textbf{Unseen}} & \multicolumn{3}{c|}{\textbf{Geo Seen}} & \multicolumn{3}{c}{\textbf{Geo Unseen}} \\
Method & Data & Embed & \textit{max} F1 & mAP5 & mAP10 & \textit{max} F1 & mAP5 & mAP10 & \textit{max} F1 & mAP5 & mAP10 & \textit{max} F1 & mAP5 & mAP10 \\
\midrule
\textcolor{gray}{Oracle} & \textcolor{gray}{S2} & \textcolor{gray}{RemoteCLIP} & \textcolor{gray}{---} & \textcolor{gray}{.500} & \textcolor{gray}{.481} & \textcolor{gray}{---} & \textcolor{gray}{.500} & \textcolor{gray}{.463} & \textcolor{gray}{---} & \textcolor{gray}{.306} & \textcolor{gray}{.286} & \textcolor{gray}{---} & \textcolor{gray}{.429} & \textcolor{gray}{.420} \\
\textcolor{gray}{Oracle} & \textcolor{gray}{S2} & \textcolor{gray}{RS-SigLIP} & \textcolor{gray}{---} & \textcolor{gray}{.489} & \textcolor{gray}{.452} & \textcolor{gray}{---} & \textcolor{gray}{.700} & \textcolor{gray}{.641} & \textcolor{gray}{---} & \textcolor{gray}{.333} & \textcolor{gray}{.313} & \textcolor{gray}{---} & \textcolor{gray}{.714} & \textcolor{gray}{.673} \\
\textcolor{gray}{Oracle} & \textcolor{gray}{PS} & \textcolor{gray}{RemoteCLIP} & \textcolor{gray}{---} & \textcolor{gray}{.565} & \textcolor{gray}{.537} & \textcolor{gray}{---} & \textcolor{gray}{.700} & \textcolor{gray}{.680} & \textcolor{gray}{---} & \textcolor{gray}{.317} & \textcolor{gray}{.286} & \textcolor{gray}{---} & \textcolor{gray}{.571} & \textcolor{gray}{.559} \\
\textcolor{gray}{Oracle} & \textcolor{gray}{PS} & \textcolor{gray}{RS-SigLIP} & \textcolor{gray}{---} & \textcolor{gray}{.715} & \textcolor{gray}{.677} & \textcolor{gray}{---} & \textcolor{gray}{.800} & \textcolor{gray}{.802} & \textcolor{gray}{---} & \textcolor{gray}{.656} & \textcolor{gray}{.595} & \textcolor{gray}{---} & \textcolor{gray}{.714} & \textcolor{gray}{.717} \\
Static & S2 & RS-SigLIP & --- & .154 & .125 & --- & .333 & .275 & --- & .096 & .079 & --- & .071 & .042 \\
Static & PS & RS-SigLIP & --- & .155 & .142 & --- & .389 & .359 & --- & .126 & .089 & --- & .000 & .002 \\
\midrule
TeoChat & --- & --- & .293 & .005 & .006 & .215 & .011 & .014 & .225 & .027 & .015 & .343 & .057 & .036 \\
SAT-JEPA-DIFF & --- & --- & --- & .035 & .025 & --- & .133 & .109 & --- & .014 & .008 & --- & .000 & .000 \\
Gemini 3 Flash & --- & --- & .482 & .110 & .086 & .273 & .309 & .266 & --- & .140 & .085 & --- & .076 & .080 \\
\midrule
Ours & S2 & RemoteCLIP & .421 & .385 & .369 & .174 & .179 & .164 & .287 & .027 & .039 & .151 & .286 & .275 \\
Ours & S2 & RS-SigLIP & .499 & .148 & .158 & .222 & .254 & .266 & .292 & .071 & .071 & .153 & .143 & .129 \\
Ours & PS & RemoteCLIP & .656 & .503 & .466 & .389 & \textbf{.520} & \textbf{.433} & .473 & .166 & .133 & .355 & .363 & .344 \\

Ours (w/o vel.) & PS & RS-SigLIP & \textbf{.954} & .511 & .488 & \textbf{.878} & .400 & .382 & \textbf{.927} & .256 & .238 & \textbf{.881} & .286 & .271 \\
\textbf{Ours} & \textbf{PS} & \textbf{RS-SigLIP} & .935 & \textbf{.533} & \textbf{.515} & .859 & .400 & .380 & .919 & \textbf{.348} & \textbf{.323} & .868 & \textbf{.429} & \textbf{.386} \\
\bottomrule
\end{tabular}%
}
\caption{\textbf{Change Detection and Event Anticipation Results.} Evaluated on 43{,}533 test samples. Seen: 54 event classes, Unseen: 10 held-out classes, Geo: geographic holdout regions. Oracle is the upper bound with perfect change detection. Our model with PlanetScope(PS) imagery and RS-SigLIP embeddings achieves the strongest retrieval performance across all settings compared against our architecture using RemoteCLIP and Sentinel-2 (S2) variations.}
\label{tab:Retrieval}
\end{table*}

%% file: tables/leadtime.tex
\centering
\resizebox{\columnwidth}{!}{%
\begin{tabular}{lccccc}
\toprule
Method & 1w--1mo & 1--3mo & 3--6mo & 6--9mo & 9mo+ \\ \midrule
\textcolor{gray}{Oracle (PS, RS-SigLIP)} & \textcolor{gray}{.522} & \textcolor{gray}{.582} & \textcolor{gray}{.603} & \textcolor{gray}{.542} & \textcolor{gray}{.472} \\
\textcolor{gray}{Oracle (PS, RemoteCLIP)} & \textcolor{gray}{.483} & \textcolor{gray}{.536} & \textcolor{gray}{.517} & \textcolor{gray}{.448} & \textcolor{gray}{.239} \\
Static (PS, RS-SigLIP) & .113 & .085 & .053 & .039 & .014 \\
Static (PS, RemoteCLIP) & .239 & .247 & .166 & .111 & .024 \\
TeoChat & .000 & .008 & .004 & .002 & .001 \\
SAT-JEPA-DIFF & .024 & .014 & .006 & .005 & .001 \\
Gemini 3 Flash & .051 & .060 & .048 & .029 & .008 \\
Ours (PS, RemoteCLIP) & .275 & .300 & .286 & .244 & .144 \\

\textbf{Ours } & \textbf{.342} & \textbf{.351} & \textbf{.354} & \textbf{.330} & \textbf{.245} \\ \bottomrule
\end{tabular}
}
\caption{\textbf{Retrieval Performance by Lead Time (mAP@10).} Our model maintains strong anticipation even at 9+ month horizons where all baselines collapse.}
\label{tab:lead_time}

%% file: tables/timetocompletion.tex
\centering
\resizebox{\columnwidth}{!}{%
\begin{tabular}{lcccccccc}
\toprule
Method & $r$ & Mean & Med. & $\leq$7d & $\leq$14d & $\leq$30d & $\leq$60d \\ \midrule
TeoChat & $-$0.133 & 97d & 80d & 11.1\% & 16.9\% & 30.9\% & 58.7\% \\
Gemini 3 Flash & $-$0.308 & 122d & 102d & 6.7\% & 9.9\% & 20.6\% & 31.9\% \\
\midrule
Ours (S2, RemoteCLIP) & 0.970 & 17d & 8d & 44.3\% & 68.4\% & 83.5\% & 93.0\% \\
Ours (S2, RS-SigLIP) & 0.992 & 10d & 6d & 54.7\% & 75.4\% & 94.8\% & 99.7\% \\
Ours (PS, RemoteCLIP) & 0.996 & 7d & 5d & 72.1\% & 88.7\% & 99.0\% & 99.9\% \\

\textbf{Ours (PS, RS-SigLIP)} & \textbf{0.997} & \textbf{5d} & \textbf{3d} & \textbf{76.8\%} & \textbf{92.1\%} & \textbf{98.9\%} & \textbf{99.9\%} \\ \bottomrule
\end{tabular}%
}
\caption{\textbf{Time-to-Completion Prediction.} Pearson correlation ($r$), mean/median absolute error, and percentage within day-error windows. Our model predicts completion dates with median 3-day error and $r{=}0.997$.}
\label{tab:time_completion}

%% file: sections/Conclusion.tex
\section{Conclusion}

For the first time in history, we are capturing a daily snapshots of the entire Earth. With this abundance of information, we want to know \textit{where}, \textit{what}, and \textit{when} global events of interest will occur. We achieve this through a novel data engine that uses vector arithmetic on VLM embeddings to discover change events at scale. This produced the largest and most diverse satellite event dataset to date featuring over 21,000 locations and 64 event types across six continents. This dataset enables us to build a first-of-its-kind global scale event anticipation model, capable of \textbf{event detection} with 93.5\% max F1 score, outperforms baselines 3.6$\times$ on \textbf{predicted event retrieval}, and \textbf{predicts completion }dates with a 3-day median error-even nine months before completion. Our automated data pipeline lowers the barrier for future work in satellite event understanding and our event model shifts Earth observation from reactive detection to proactive anticipation of the events that shape our world.


%% file: sections/Appendix.tex
\begin{figure*}[ht]
    \centering
    \includegraphics[width=0.9\linewidth]{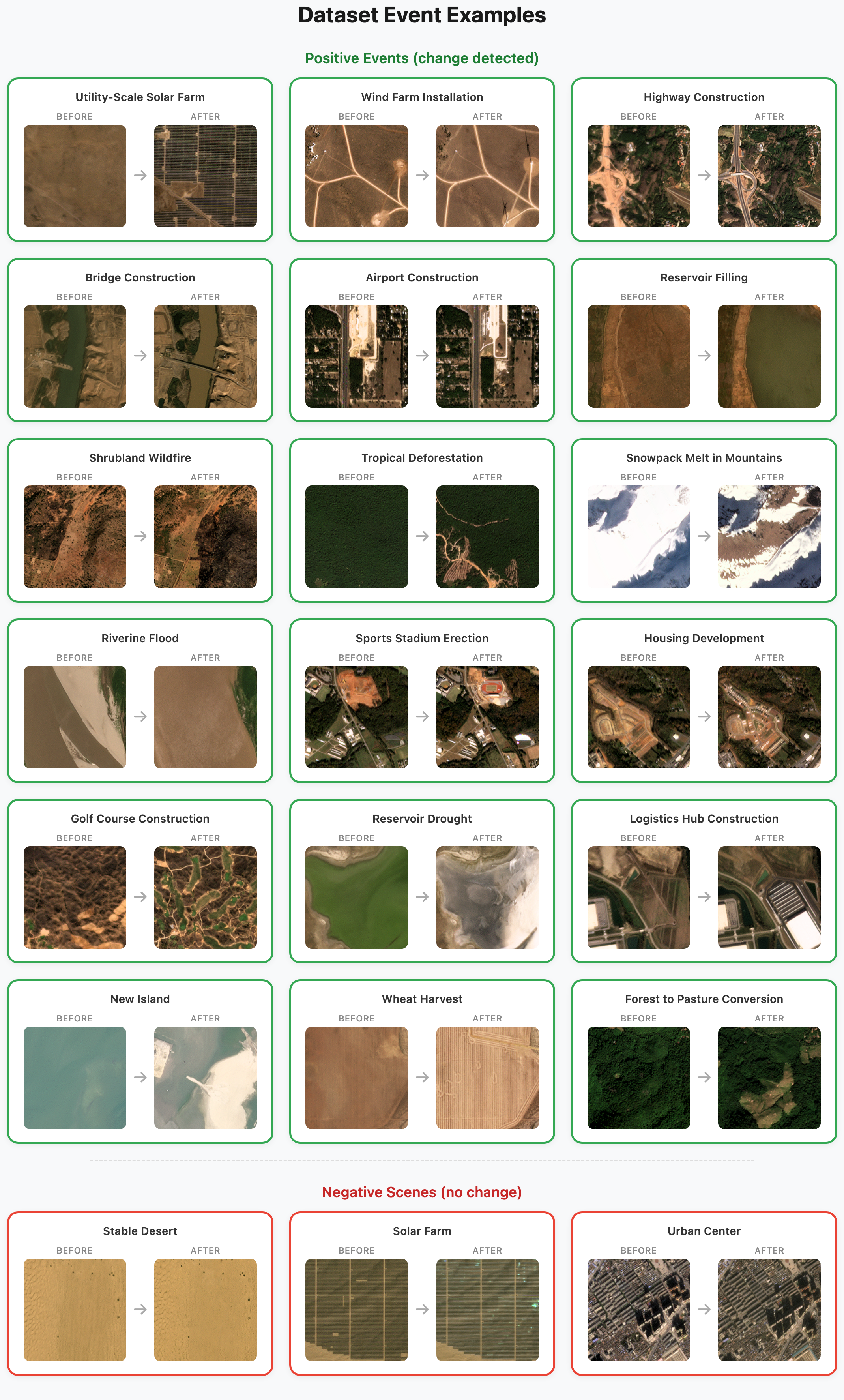}
    \caption{\small \textbf{Dataset Event Examples.} 
    Representative before and after image pairs from our automatically-discovered dataset. 
    \textit{Top (green):} Positive events showing diverse changes. 
    \textit{Bottom (red):} Negative scenes showing stable locations with no significant change over time. In total we have 64 positive event classes and 69 stable scene classes.}
    \label{fig:dataset_examples}
\end{figure*}

\begin{figure}[p]                               
      \centering                                                
      \includegraphics[width=\textwidth]{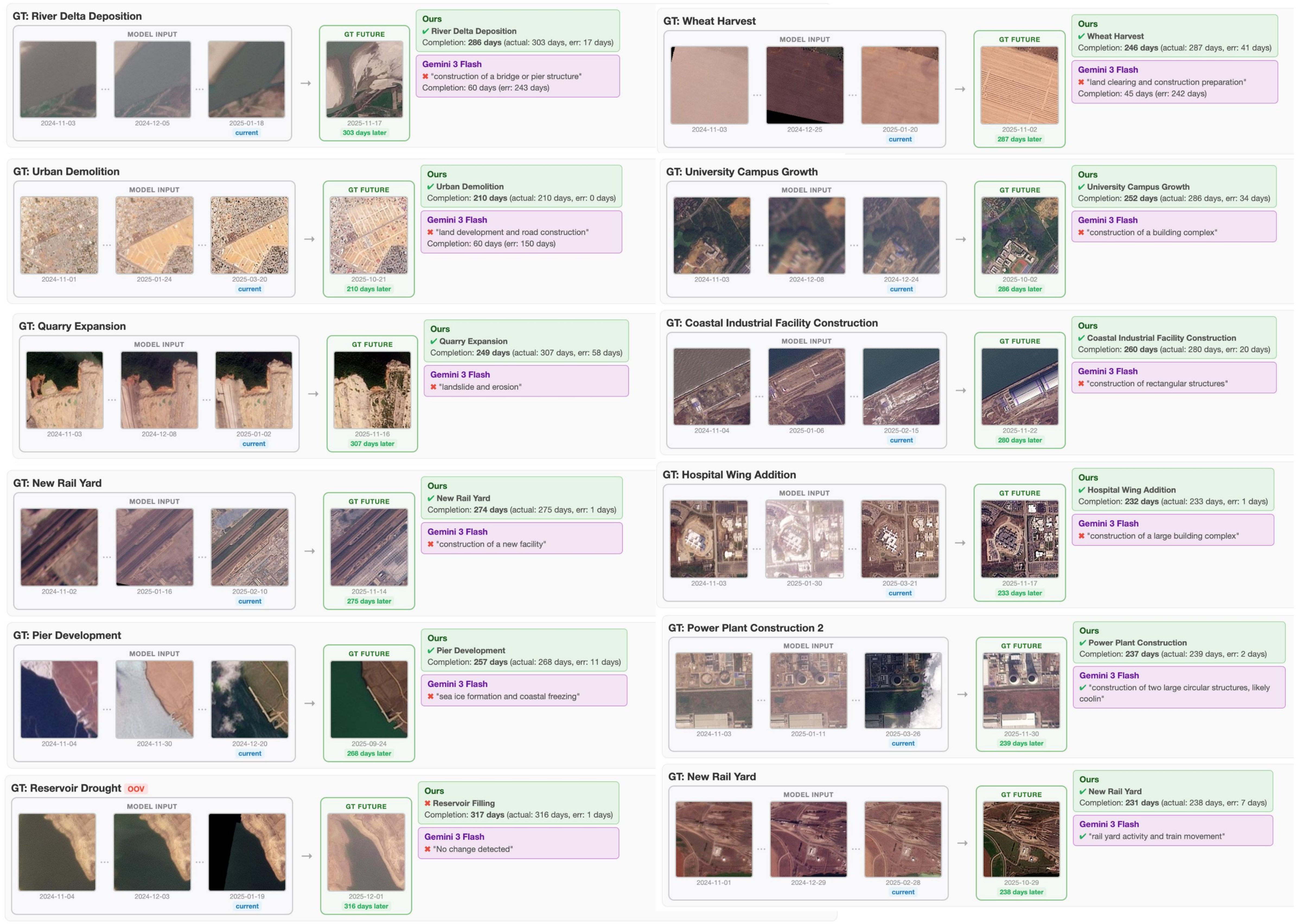}
      \caption{Qualitative prediction examples. Each card shows samples from the multi-temporal input sequences as well as the future state of the event with days until completion. To the right we include our model's prediction vs Gemini 3 Flash}
      \label{fig:append_qualitative}
  \end{figure}

\begin{figure}
    \centering
    \includegraphics[width=\linewidth]{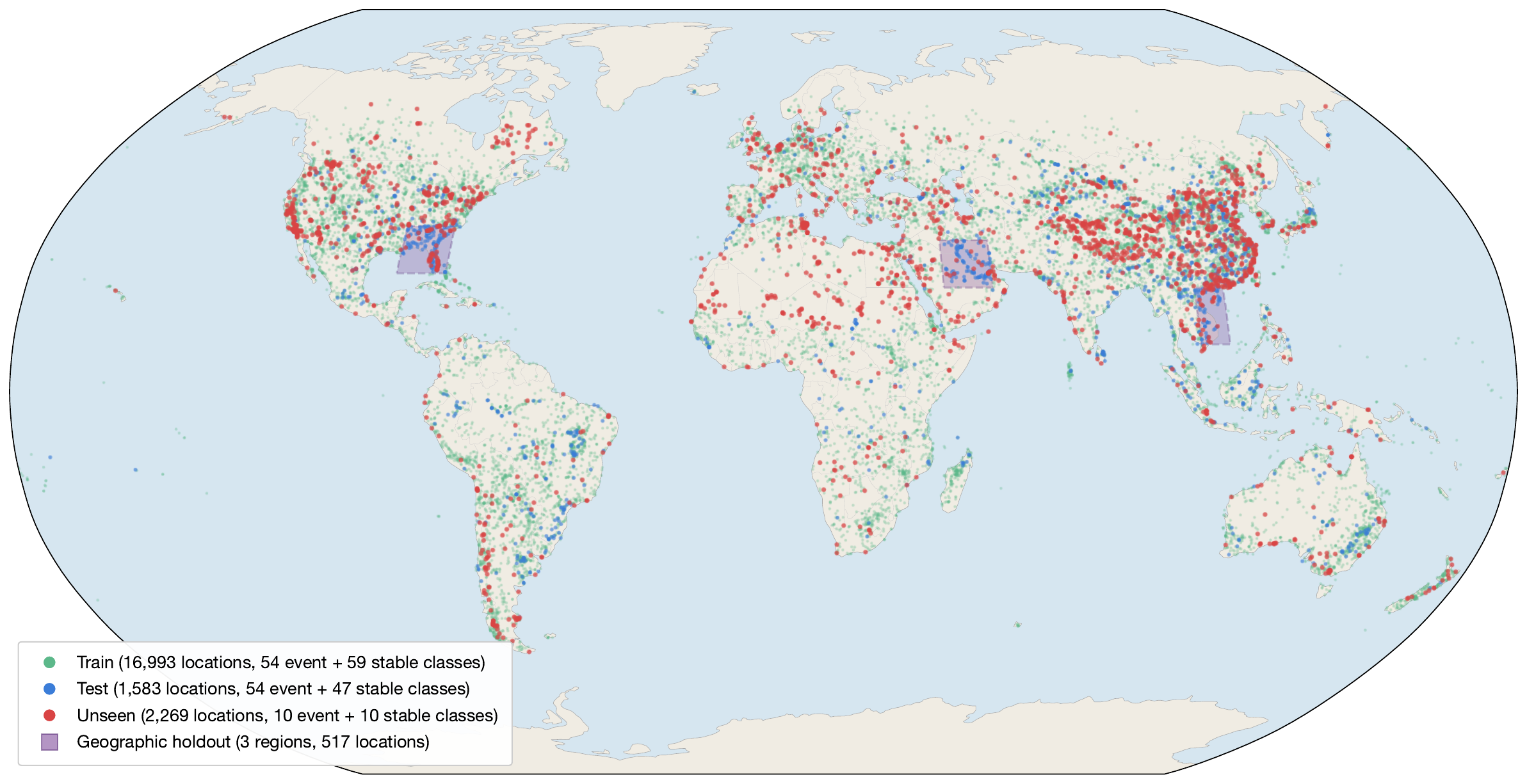}
    \caption{\textbf{Dataset Geographic Distribution.} Our global event dataset comprises over 21,000+ locations. We explicitly partition the data into a training set (green), test set (blue), and an unseen class set (red) to evaluate zero-shot event anticipation, and three geographic holdout regions (purple) in North America, the Middle East, and East Asia to test regional robustness.}
    \label{fig:dataset_dist}
\end{figure}

  \begin{figure*}[ht]
    \centering
    \includegraphics[width=\linewidth]{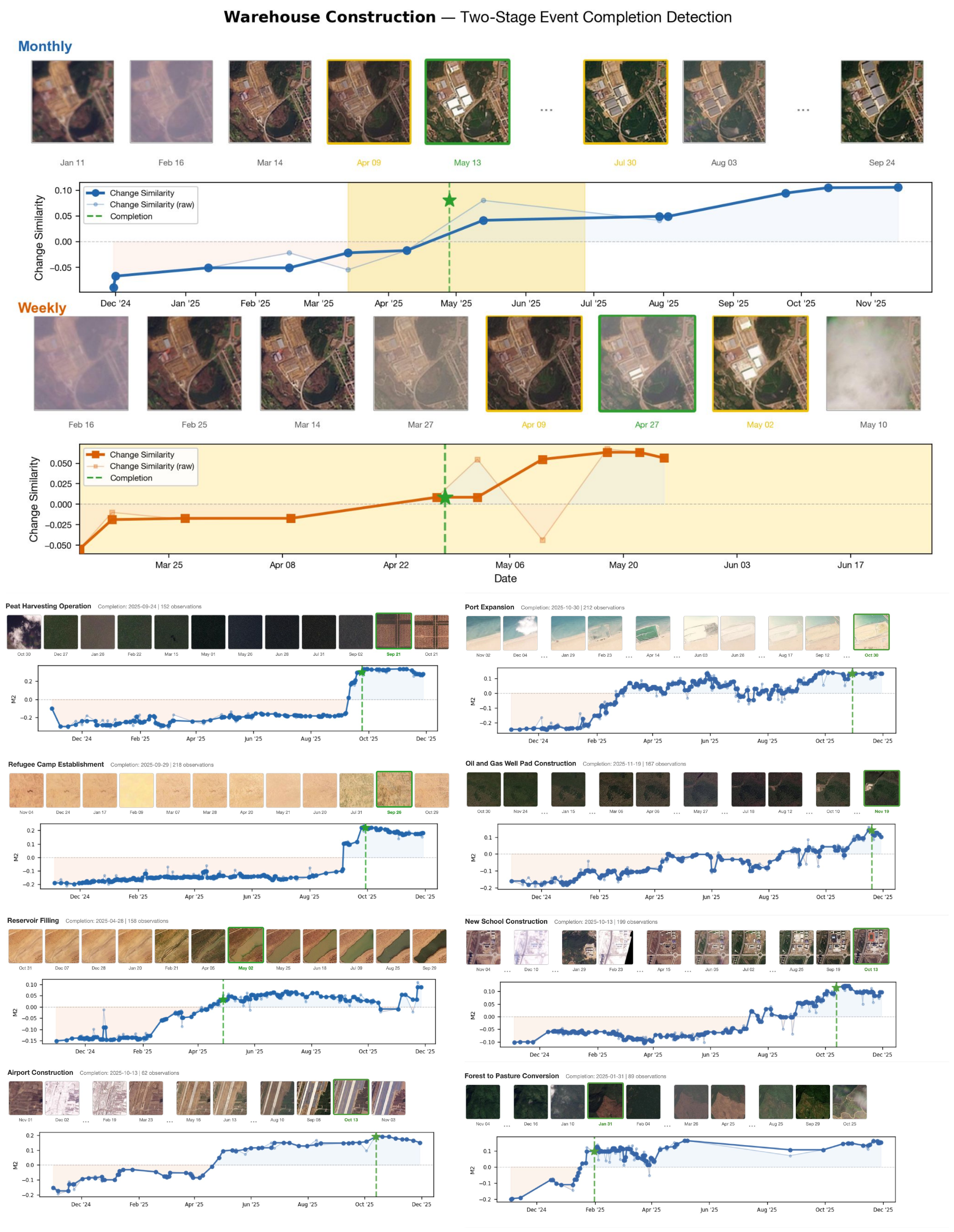}
    \caption{\small \textbf{Event Completion Detection.} We demonstrate our two-phased completion detection process for a \textit{Warehouse Construction} event. The visualization shows the transition from a monthly to a weekly analysis level to increase temporal resolution. Section.~\ref{subsubsec:event_completion_detection} and Appendix~\ref{appendix:completion_detection}. Below, additional examples illustrate completion detection across diverse events. }
    \label{fig:completion_detection}
\end{figure*}

\begin{figure}[p]
    \centering                         \includegraphics[width=0.6\linewidth]{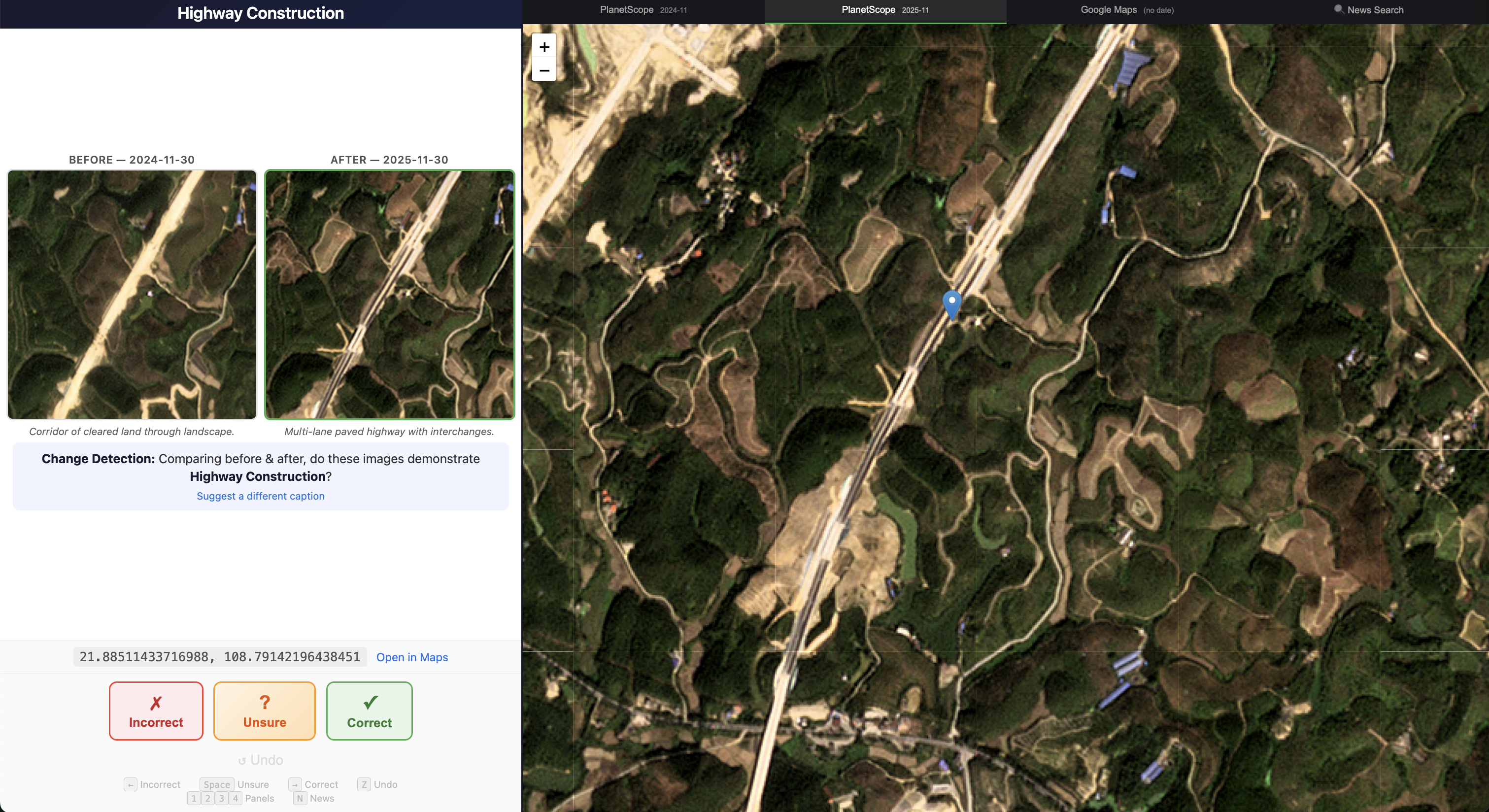}                             \includegraphics[width=0.6\linewidth]{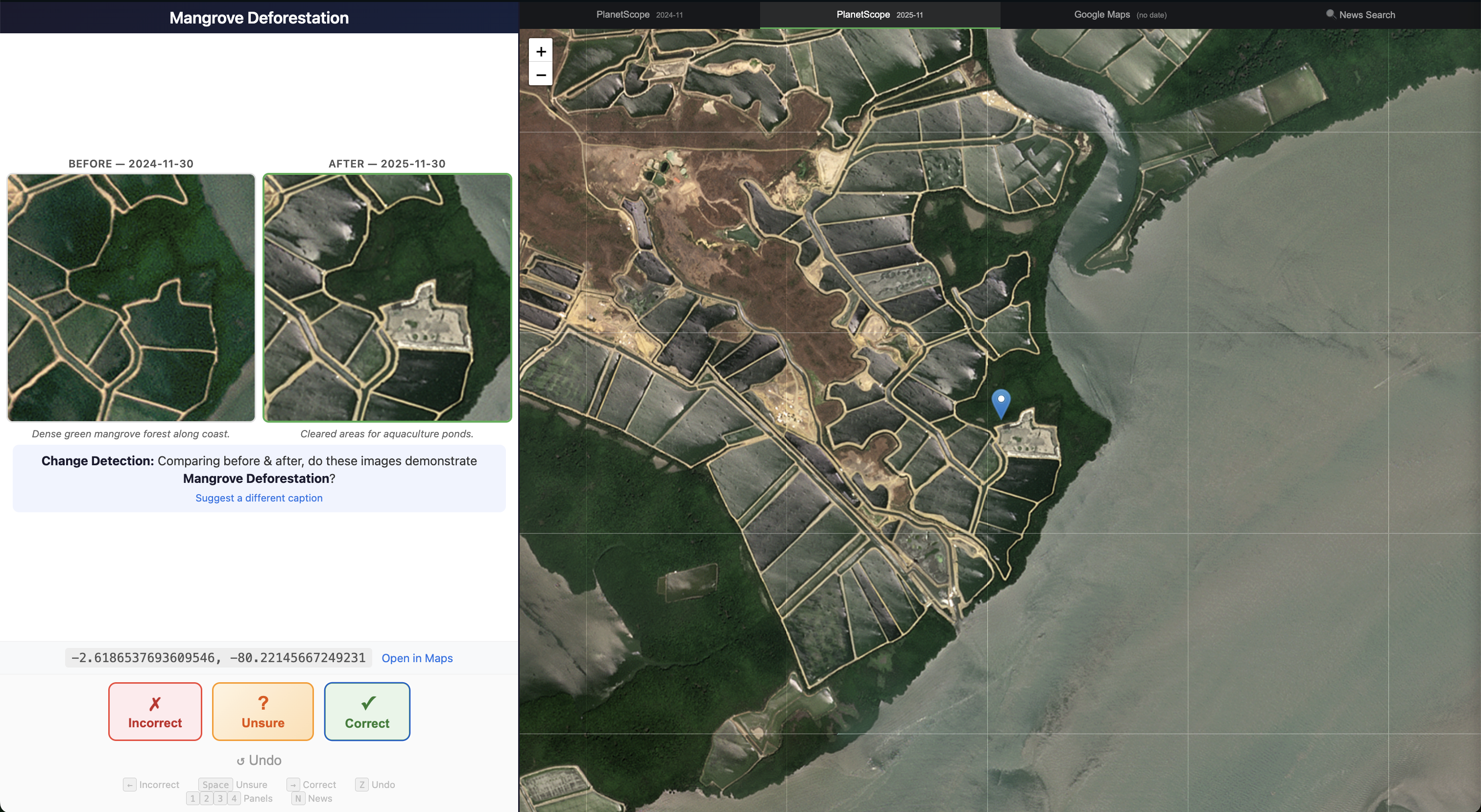}                             \includegraphics[width=0.6\linewidth]
    {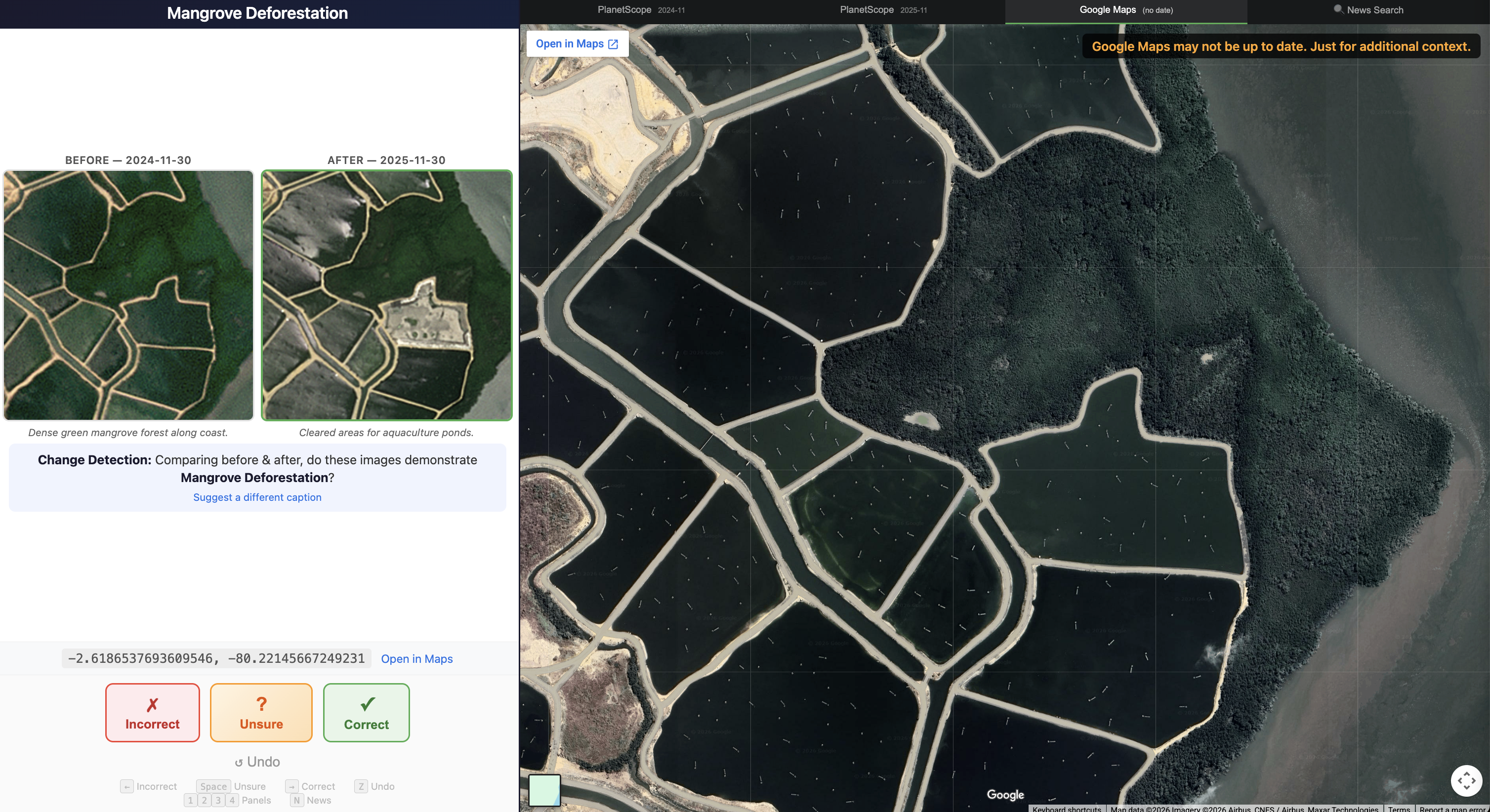}
     \includegraphics[width=0.6\linewidth]{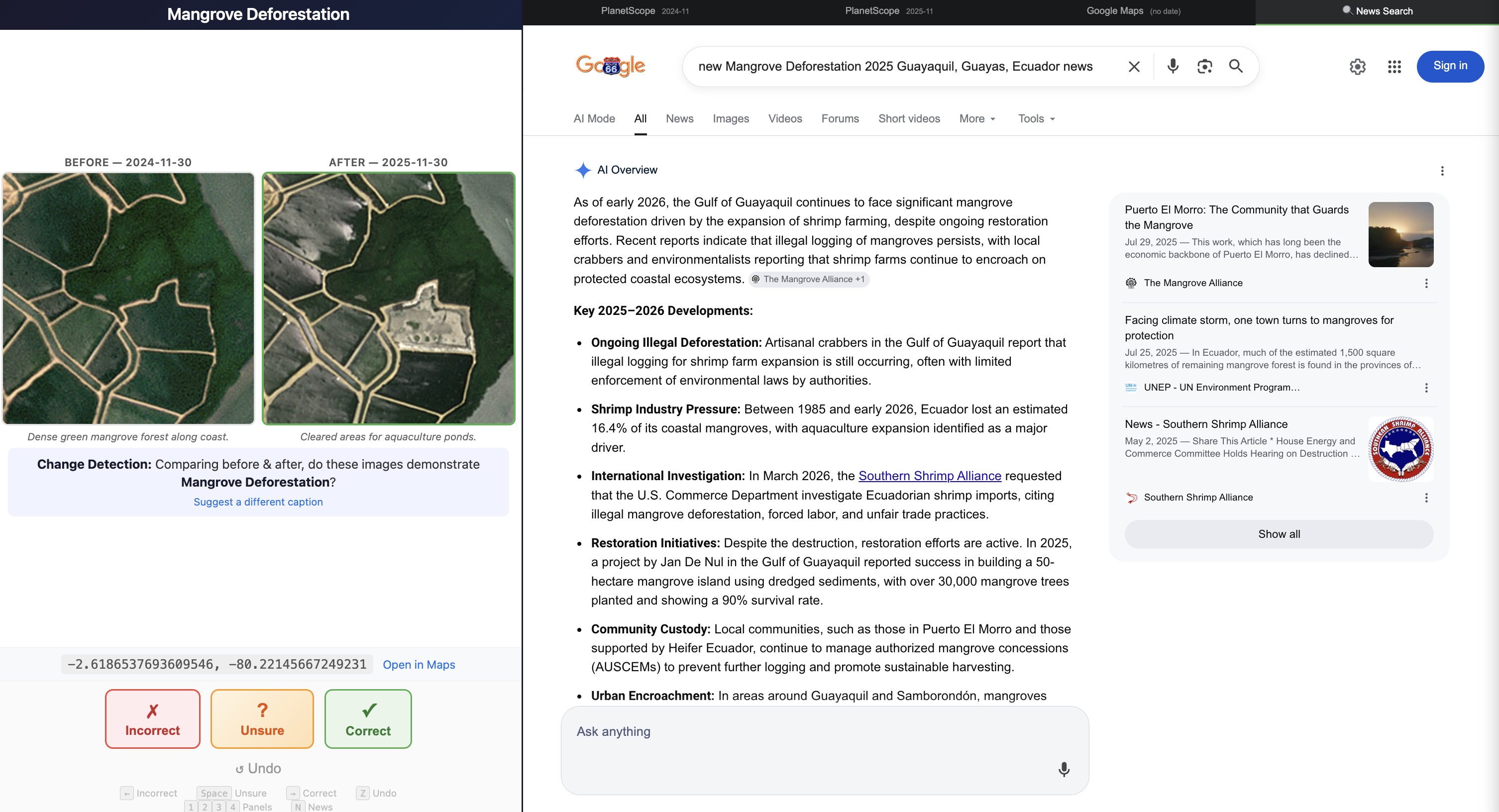} 
    
    \caption{\textbf{Screenshots of the human evaluation interface.} The top row shows two example events, \textit{highway construction} and \textit{mangrove deforestation}, followed by the corresponding maps and news results for the mangrove example for reference. Evaluators used these 4 context pages to label the event with "correct", "incorrect" or "unsure". Sampling with multiple "correct" labels were used to create our test sets.}
    \label{fig:human-eval}
  \end{figure}

\clearpage  
\appendix

\section{Examples of Satellite Event Dataset}

We include some examples of the events from our dataset in Figure~\ref{fig:dataset_examples} showcasing its diversity.

\section{Additional Qualitative Results of Predicted Event Retrieval}
\label{appendix:qualitative_overall}
We visualize a variety of qualitative results from our test set, starting with samples where our model predicts events several months before completion, as well as samples where our models fall short in Figure~\ref{fig:append_qualitative}.
Each sample shows the input sequences followed by the future state to be predicted.

\section{Model Comparison for Event Retrieval}
\label{appendix:event_retrieval}
\input{tables/event_retrieval}
We evaluate several popular satellite VLMs for event retrieval to determine which models we should use to train our world model. From Table.~\ref{tab:event_retrieval_dataset}, one can see that the performance of RS-Siglip was much higher than the open source models. For this reason we finetuned RemoteCLIP to ensure that we had an opensource model that was able to recognize the \textit{before} and \textit{after} states of our events.

While this improved its understanding visually of the event, we see that it improved the \textit{Intersection} retrieval far more than the \textit{Change Vector} retrieval.To better test the change understanding of the models, we calculate the \textit{Intra-Event Consistency}, asking For events with 2+ samples, how similar are their change vectors? Here we calculate the mean pairwise cosine similarity between change vectors of the same event, from this we found that the RS-Siglip model held a 0.5606 vs the finetuned RemoteCLIP models carried a 0.4695 and off-the-shelf RemoteCLIP held 0.3121.   

The fine-tuned RemoteCLIP learned better representations of the individual states (before/after), which helped it find relevant images via intersection retrieval. But when you look at the direction of change (after minus before), RS-SigLIP's change vectors for the same event type are more tightly aligned with each other. We hypothesize that this consistency, i.e. the degree to which a model produces a stable change direction for a given event type directly affects prediction quality. Supporting this, Table ~\ref{tab:Retrieval} shows that RS-SigLIP embeddings yield stronger prediction performance, particularly on unseen events (mAP5 .254 vs .200), where consistent change geometry would matter most for generalization.

\section{Ablation Study on Anticipation Model Architectures}
\label{appendix:architecture_ablation}
To test whether the anticipation gains depend on the transformer backbone,
we replace it with an LSTM and a GRU while keeping the inputs
(velocity-augmented embeddings), horizon conditioning, training objective,
and data fixed. All three architectures outperform the Static baseline at each $k$.

\begin{table}[th]
\centering
\resizebox{0.6\columnwidth}{!}{%
\begin{tabular}{lcccc}
\toprule
Method & max F1 & mAP@5 & mAP@10 & mAP@20 \\ \midrule
\textcolor{gray}{Oracle (PS, RS-SigLIP)} & \textcolor{gray}{--} & \textcolor{gray}{.734} & \textcolor{gray}{.705} & \textcolor{gray}{.642} \\
Static (PS, RS-SigLIP) & -- & .208 & .191 & .153 \\
TeoChat & .275 & .005 & .006 & .006 \\
SAT-JEPA-DIFF & -- & .057 & .044 & .032 \\
Gemini 3 Flash & .435 & .155 & .126 & .100 \\
Ours (LSTM) & .766 & .361 & .315 & .270 \\
Ours (GRU) & .839 & .401 & .348 & .269 \\
\textbf{Ours (Transformer)} & \textbf{.949} & \textbf{.503} & \textbf{.485} & \textbf{.415} \\ \bottomrule
\end{tabular}
}
\caption{\textbf{Anticipation Architecture Ablation.} Evaluated over the seen and unseen event classes combined. All three backbones outperform every baseline, and the transformer is best on every metric.}
\label{tab:arch_ablation}
\end{table}

\section{Global Distribution of Event Dataset}
\label{appendix:dataset_map}
Figure.~\ref{fig:dataset_dist} visualizes the global distribution of our event dataset, as well as the locations of the test set, unseen classes, and geographic hold out regions.

\section{Implementation Details of Event Completion Detection}
\label{appendix:completion_detection}
\textbf{Noise filtering.} Before analyzing the $S_{\Delta}$ curve, we detect and exclude cloudy or noisy images. For each embedding, we compute cosine similarity to both temporal neighbors and set an adaptive threshold based on the median neighbor similarity across all images. Images with low similarity to both neighbors are flagged as outliers and excluded from subsequent analysis. We then apply a median filter to the $S_{\Delta}$ series to suppress remaining noise while preserving endpoints.

\textbf{Coarse-to-fine completion detection.} At monthly resolution, we detect the first sustained zero-crossing---the earliest timepoint where the signal transitions from negative to positive and remains positive for at least $k$ subsequent observations. Small epsilon margins ($\epsilon^-$, $\epsilon^+$) avoid false detections from noise. If no clear crossing exists, we fall back to the temporal window with the highest mean change similarity, and finally to the maximum first-order difference.

To refine from $\sim$30-day to $\sim$7-day precision, we download weekly imagery ($\sim$1 image/week) for the detected completion month plus one month before and after. We repeat the noise filtering and zero-crossing analysis on this weekly $S_{\Delta}$ series, yielding a precise completion estimate to the week. We visualize this process for a warehouse construction event in Figure~\ref{fig:completion_detection}. The completion timestamps are a crucial part of a temporal event dataset and these annotations are included for each event in the dataset. 

As shown in Figure~\ref{fig:completion_detection}, the $S_{\Delta}$ signal exhibits sharp zero-crossings across diverse event types, confirming that our completion dates capture genuine perceptual transitions rather than gradual drift.

\section{Human Verification of Test Data}
\label{appendix:human_verification_app}
Screenshots from our web-based labeling interface where annotators classify satellite before/after image pairs as showing the described change event or not. Annotators get to see beyond the boundaries of the single image for both time points, as well as map and search results for the event type in the area for additional context.

\section{Implementation Details of Anticipation Model }
\label{appendix:model_implementation_details}

Our model uses a 4-layer causal transformer with 8 attention heads, feed-forward dimension 2048, and dropout 0.2. Input sequences are padded to a maximum of 53 tokens. We train with AdamW ($\text{lr}=10^{-4}$, weight decay $10^{-3}$) and cosine annealing for 200 epochs with batch size 32 and gradient clipping at 1.0. The magnitude regularization weight is $\lambda = 0.1$. Model size scales with the VLM backbone: 54.8M parameters for RS-SigLIP (1152-dim) and 29.7M for RemoteCLIP (768-dim). All models are trained on a single NVIDIA A100 40GB GPU with 170GB system memory. Training the PS RS-SigLIP model (177K samples, 200 epochs) takes approximately 14 hours; S2 variants (~163K samples) train in approximately 12 hours. We select the best checkpoint by seen-class mAP@10 on the validation split. All experiments use the same hyperparameters with no per-backbone tuning.

\section{Logging Permit Dataset}
\label{appendix:logging_permit}

In Section~\ref{sec:results} we discuss one of the limitations of our work which is the ambiguity of time-to-completion. We created an independent, unbiased test set by labeling a small set of logging and deforestation events recorded in publicly available Washington State DNR logging-permit records~\cite{logging_permit}, so that no sample was selected using embeddings. Each site has hand-verified dates for when harvesting first becomes visible and when it is complete. We run two experiments on this data.

\textbf{(1)} How accurate is our embedding-based completion detector on events not collected via embeddings? 
On 14 hand-labeled sites we found that our embeddings driven completion detection agreed with the human labels with a median of 8 days (79\% within 30 days), with two outliers at sites where the clearing happened in several stages.

\textbf{(2)} How well do we detect and retrieve these events? We ranked every chip in an 8,878 km² region of Washington by predicted logging, 2–4 months ahead. On visual inspection, 7 of the model's top 20 picks were confirmed future clear-cuts (static baseline: 3), and only 1 pick showed no logging at all.

\section{Time-to-completion Ambiguity}
\label{appendix:timetocompletionambiguity}
Event completion is ambiguous, especially when observed remotely. We define completion as the point at which a location becomes retrievable as its ``after'' state, but this can disagree with a human assessment of \textit{completion}. For example, when a data center is constructed, the main building may appear finished from above months before the parking lots and auxiliary facilities around it are complete, so different observers may reasonably place completion at different times.

To assess this disagreement, we compare human-labeled completion dates with our embedding-based completion detection. Agreement varies by event type: where the end state is visually clear (housing, highways, bridges), the two match closely (median 8--24 days); where the change is gradual, cyclical or reversible, and humans themselves find completion hard to pin down, they disagree more.

Completion is therefore not a single well-defined moment for every event type. This supports our definition, which applies consistently across event types and agrees closely with human judgement where the end state is clear.

%% file: tables/event_retrieval.tex
\begin{table}[h]
\centering
\resizebox{0.7\columnwidth}{!}{%
\begin{tabular}{lccc|ccc}
\toprule
& \multicolumn{3}{c|}{\textbf{Change mAP@K}} & \multicolumn{3}{c}{\textbf{Intersection mAP@K}} \\
Method & @5 & @10 & @20 & @5 & @10 & @20 \\ \midrule
RemoteCLIP FT & \textbf{0.752} & \textbf{0.760} & \textbf{0.764} & \textbf{0.822} & \textbf{0.835} & \textbf{0.840} \\
RS-SigLIP~\cite{barzilai2025recipe} & 0.688 & 0.709 & 0.710 & 0.691 & 0.716 & 0.720 \\
GIT-RS-CLIP~\cite{Text2Earth} & 0.400 & 0.418 & 0.431 & 0.422 & 0.459 & 0.478 \\
DOFA-CLIP-B~\cite{xiong2025dofa} & 0.318 & 0.338 & 0.356 & 0.350 & 0.380 & 0.396 \\
RemoteCLIP~\cite{liu2024remoteclip} & 0.262 & 0.287 & 0.294 & 0.327 & 0.365 & 0.380 \\ \bottomrule
\end{tabular}%
}
\caption{\textbf{Zero-shot Event Retrieval.} Comparative evaluation of satellite VLMs performance using Intersection Search and Change Vector Search as described in Sec.~\ref{sec:dataset}.}
\label{tab:event_retrieval_dataset}
\end{table}

%% file: checklist.tex
\section*{NeurIPS Paper Checklist}

\begin{enumerate}

\item {\bf Claims}
    \item[] Question: Do the main claims made in the abstract and introduction accurately reflect the paper's contributions and scope?
    \item[] Answer: \answerYes{} 
    \item[] Justification: Yes, we claim to present a new global event retrieval method, and a new dataset with this pipeline, followed by an event prediction model. Which we present results for in the results section.

\item {\bf Limitations}
    \item[] Question: Does the paper discuss the limitations of the work performed by the authors?
    \item[] Answer:  \answerYes{} 
    \item[] Justification: We include a limitations section in our discussion in Section.~\ref{sec:results}.

\item {\bf Theory assumptions and proofs}
    \item[] Question: For each theoretical result, does the paper provide the full set of assumptions and a complete (and correct) proof?
    \item[] Answer:\answerNA{} 
    \item[] Justification: We present capabilities demonstrated by embeddings from VLMs, and mathematically describe them, however we do not have any theoretical proofs. 

    \item {\bf Experimental result reproducibility}
    \item[] Question: Does the paper fully disclose all the information needed to reproduce the main experimental results of the paper to the extent that it affects the main claims and/or conclusions of the paper (regardless of whether the code and data are provided or not)?
    \item[] Answer: \answerYes{} 
    \item[] Justification: All code and data will be publicly released and we include implementation details in Section~\ref{sec:experiments} and Appendices~\ref{appendix:completion_detection},~\ref{appendix:human_verification_app} and ~\ref{appendix:model_implementation_details}.

\item {\bf Open access to data and code}
    \item[] Question: Does the paper provide open access to the data and code, with sufficient instructions to faithfully reproduce the main experimental results, as described in supplemental material?
    \item[] Answer: \answerYes{} 
    \item[] Justification: All code and data will be publicly released upon publication.

\item {\bf Experimental setting/details}
    \item[] Question: Does the paper specify all the training and test details (e.g., data splits, hyperparameters, how they were chosen, type of optimizer) necessary to understand the results?
    \item[] Answer: \answerYes{} 
    \item[] Justification: We include train and test details in Section.~\ref{sec:experiments} and model implementation details in Appendix.~\ref{appendix:model_implementation_details}.

\item {\bf Experiment statistical significance}
    \item[] Question: Does the paper report error bars suitably and correctly defined or other appropriate information about the statistical significance of the experiments?
    \item[] Answer: \answerYes{}
    \item[] Justification: To test robustness and statistical significance of our results we perform a bootstrapping experiment by subsampling within our test set 1,000 times, and a permutation test on the per-class AP between the static baseline and our model using 10,000 permutations.

\item {\bf Experiments compute resources}
    \item[] Question: For each experiment, does the paper provide sufficient information on the computer resources (type of compute workers, memory, time of execution) needed to reproduce the experiments?
    \item[] Answer: \answerYes{} 
    \item[] Justification: Yes we include compute resources used in our implementation details in Appendix~\ref{appendix:model_implementation_details}.
    
\item {\bf Code of ethics}
    \item[] Question: Does the research conducted in the paper conform, in every respect, with the NeurIPS Code of Ethics \url{https://neurips.cc/public/EthicsGuidelines}?
    \item[] Answer: \answerYes{} 
    \item[] Justification: We conform to the NeurIPS Code of Ethics.

\item {\bf Broader impacts}
    \item[] Question: Does the paper discuss both potential positive societal impacts and negative societal impacts of the work performed?
    \item[] Answer: \answerYes{} 
    \item[] Justification: We include a paragraph on the broader impacts of our work in Section.~\ref{sec:results}, paragraph~\ref{para:broader_impact}.
    
\item {\bf Safeguards}
    \item[] Question: Does the paper describe safeguards that have been put in place for responsible release of data or models that have a high risk for misuse (e.g., pre-trained language models, image generators, or scraped datasets)?
    \item[] Answer: \answerYes{} 
    \item[] Justification: We do not include any sensitive information in our dataset.

\item {\bf Licenses for existing assets}
    \item[] Question: Are the creators or original owners of assets (e.g., code, data, models), used in the paper, properly credited and are the license and terms of use explicitly mentioned and properly respected?
    \item[] Answer: \answerYes{}{} 
    \item[] Justification: The creators and original owners are cited and credited. We use two commercially available products, i.e. the Rs-sigLIP model and Planetscope imagery both of which are credited and with legal permission to publish. We include the MCLA license of Planetscope with the CC BY-SA 4.0 license applying to our dataset and code on release.

\item {\bf New assets}
    \item[] Question: Are new assets introduced in the paper well documented and is the documentation provided alongside the assets?
    \item[] Answer: \answerYes{}{} 
    \item[] Justification: We include details of the dataset and plan to release datasets and code upon publication, and have received permission to release Planetscope imagery for research purposes under CC BY-SA 4.0 licensing.

\item {\bf Crowdsourcing and research with human subjects}
    \item[] Question: For crowdsourcing experiments and research with human subjects, does the paper include the full text of instructions given to participants and screenshots, if applicable, as well as details about compensation (if any)? 
    \item[] Answer: \answerYes{} 
    \item[] Justification: Screenshots of our labelling application are including in Appendix~\ref{appendix:human_verification_app}, no compensation was provided and no personal data was collected.

\item {\bf Institutional review board (IRB) approvals or equivalent for research with human subjects}
    \item[] Question: Does the paper describe potential risks incurred by study participants, whether such risks were disclosed to the subjects, and whether Institutional Review Board (IRB) approvals (or an equivalent approval/review based on the requirements of your country or institution) were obtained?
    \item[] Answer: \answerNA{} 
    \item[] Justification:The human verification was conducted  on a voluntary basis, no personal data was collected, and no compensation was provided. The activity did not constitute human subjects research as defined by IRB guidelines, and therefore no IRB approval was required.

\item {\bf Declaration of LLM usage}
    \item[] Question: Does the paper describe the usage of LLMs if it is an important, original, or non-standard component of the core methods in this research? Note that if the LLM is used only for writing, editing, or formatting purposes and does \emph{not} impact the core methodology, scientific rigor, or originality of the research, declaration is not required.
    \item[] Answer: \answerYes{}{} 
    \item[] Justification: We describe the usage of MLLMs in the ARF and Event description generation components of our work in Sections~\ref{sec:dataset} and ~\ref{sec:experiments}.

\end{enumerate}